\RequirePackage{fix-cm}
\documentclass[twocolumn]{svjour3}          
\smartqed  
\usepackage{upgreek}
\usepackage{acronym}

\usepackage[table]{xcolor}
\usepackage{amsmath}
\usepackage{amssymb}
\usepackage{mathrsfs}
\usepackage{natbib}
\usepackage{array}
\usepackage{graphicx}
\usepackage[
            pdfpagelabels,hypertexnames=true,
            plainpages=false,
            naturalnames=false]{hyperref}

\usepackage{color}
\definecolor{darkblue}{rgb}{0,0,1}
\hypersetup{colorlinks,
            linkcolor=darkblue,
            anchorcolor=darkblue,
            citecolor=darkblue}
\usepackage{rotating}
\usepackage{booktabs}
\usepackage{tikz}
\usepackage{subfigure}
\usepackage{multicol}
\usepackage{multirow}
\usepackage{rotating}
\usepackage{arydshln}
\usepackage{floatrow}

\usepackage{enumerate}
\usepackage{algorithmic}
\usepackage[lined,ruled,linesnumbered]{algorithm2e}
\usepackage{microtype}
\usepackage{booktabs}
\usepackage{float}

\usepackage{times}
\usepackage{epsfig}
\usepackage{graphicx}
\usepackage{color}
\usepackage{subfigure}
\usepackage[misc,geometry]{ifsym}
\usepackage{etoolbox}

\makeatletter
\def\hlinewd#1{%
  \noalign{\ifnum0=`}\fi\hrule \@height #1 \futurelet
   \reserved@a\@xhline}
\makeatother

\newcommand{\mcrot}[4]{\multicolumn{#1}{#2}{\rlap{\rotatebox{#3}{#4}~}}} 
\newcommand*{\twoelementtable}[3][l]%
{%
    \renewcommand{\arraystretch}{0.8}%
    \begin{tabular}[t]{@{}#1@{}}%
        #2\tabularnewline
        #3%
    \end{tabular}%
}
\newfloatcommand{capbtabbox}{table}[][\FBwidth]

\definecolor{colour1}{rgb}{0.9, 1, 1}
\definecolor{colour2}{rgb}{1, 0.9, 0.8}
\definecolor{colour3}{rgb}{0.9, 1, 0.9}
\definecolor{colour4}{rgb}{1, 0.9, 0.9}
\definecolor{colour5}{rgb}{0.95, 0.9, 1}

\definecolor{colourBlock1}{rgb}{255,153,153}
\definecolor{colourBlock2}{rgb}{255,204,153}
\definecolor{colourBlock3}{rgb}{255,255,153}
\definecolor{colourBlock4}{rgb}{153,255,153}
\definecolor{colourBlock5}{rgb}{102,255,255}
\definecolor{colourBlock6}{rgb}{102,204,255}
\definecolor{colourBlock7}{rgb}{153,153,255}
\definecolor{colourVecOp}{rgb}{127,127,127}
\definecolor{colourFC1}{rgb}{255,153,255}
\definecolor{colourFC2}{rgb}{255,153,204}
\definecolor{colourFC3}{rgb}{255,51,204}

\definecolor{mygreen}{rgb}{0,0.7,0}

\begin{document}

\title{A Comprehensive Analysis of Weakly-Supervised Semantic Segmentation in Different Image Domains
}


\author{Lyndon Chan \and Mahdi S. Hosseini \and Konstantinos N. Plataniotis}

\authorrunning{Lyndon Chan \and Mahdi S. Hosseini \and Konstantinos N. Plataniotis} 

\institute{Lyndon Chan (\Letter) \and Mahdi S. Hosseini (\Letter) \and Konstantinos N. Plataniotis
						\at The Edward S. Rogers Sr. Department of Electrical \& Computer Engineering, University of Toronto, Toronto, Canada\\\email{lyndon.chan@mail.utoronto.ca} \\ \email{mahdi.hosseini@mail.utoronto.ca}
}


\date{Received: date / Accepted: date}

\maketitle

\begin{abstract}
Recently proposed methods for weakly-supervised semantic segmentation have achieved impressive performance in predicting pixel classes despite being trained with only image labels which lack positional information. Because image annotations are cheaper and quicker to generate, weak supervision is more practical than full supervision for training segmentation algorithms. These methods have been predominantly developed to solve the background separation and partial segmentation problems presented by natural scene images and it is unclear whether they can be simply transferred to other domains with different characteristics, such as histopathology and satellite images, and still perform well. This paper evaluates state-of-the-art weakly-supervised semantic segmentation methods on natural scene, histopathology, and satellite image datasets and analyzes how to determine which method is most suitable for a given dataset. Our experiments indicate that histopathology and satellite images present a different set of problems for weakly-supervised semantic segmentation than natural scene images, such as ambiguous boundaries and class co-occurrence. Methods perform well for datasets they were developed on, but tend to perform poorly on other datasets. We present some practical techniques for these methods on unseen datasets and argue that more work is needed for a generalizable approach to weakly-supervised semantic segmentation. Our full code implementation is available on GitHub: \url{https://github.com/lyndonchan/wsss-analysis}.


\end{abstract}

\section{Introduction}\label{sec_introduction}


Multi-class semantic segmentation aims to predict a discrete semantic class for every pixel in an image. This is useful as an attention mechanism: by ignoring the irrelevant parts of the image, only relevant parts are retained for further analysis, such as faces and human parts \citep{prince2012computer1}. Semantic segmentation is also useful for changing the pixels of the image into higher-level representations that are more meaningful for further analysis, such as object locations, shapes, sizes, textures, poses, or actions \citep{shapiro2000computer}. Oftentimes, semantic segmentation is used when simply predicting a bounding box around the objects is too coarse for fine-grained tasks, especially when the scene is cluttered and the bounding boxes would overlap significantly or when the precise entity boundaries are important. Whereas humans can ordinarily perform such visual inspection tasks accurately but slowly, computers have the potential to perform the same tasks at larger scale and with greater accuracy \citep{prince2012computer2}. Natural scene images can be segmented to monitor traffic density \citep{audebert2017segment}, segment humans from images \citep{xia2017joint}, and gather crowd statistics \citep{zhang2015cross}. Histopathology images can be segmented to detect abnormally-shaped renal tissues \citep{kothari2013histological}, quantify cell size and density \citep{lenz2016estimating}, and build tissue-based image retrieval systems \citep{zhang2015fine}. Finally, satellite images can be segmented to detect weeds in farmland \citep{gao2018fusion}, detect flooded areas \citep{rahnemoonfar2018flooded}, and quantify urban development \citep{zhang2019detecting}.

The most popular approach to training semantic segmentation models is currently full supervision, whereby the ground-truth pixel segmentation map is observable for training. Fully-supervised semantic segmentation (FSS) methods include OCNet \citep{yuan2019object}, DANet \citep{fu2019dual}, HRNet \citep{wang2019deep}, FCN \citep{long2015fully}, U-Net \citep{ronneberger2015u}, sliding window DNN \citep{ciresan2012neuronal}, and multiscale convnet \cite{farabet2012learning}. Although fully-supervised methods attain state-of-the-art performance, annotating each training image by pixel is costly and slow. The labellers of MS COCO took on average 4.1 seconds to label each image by category and 10.1 minutes to label each image by pixel-level instances \citep{lin2014microsoft}, requiring 150 times the time needed for image-level annotations. Apart from full supervision, other approaches have been proposed to reduce the annotation cost: the unsupervised approach uses unlabelled images, the semi-supervised approach uses a combination of labelled and unlabelled images (or of reliable and noisily-labelled images), and the weakly-supervised approach uses less spatially-informative annotations than the pixel level. Of these approaches, weak supervision generally performs best; its training annotations (in order of decreasing informativeness) include: bounding box \citep{dai2015boxsup, papandreou2015emadapt}, scribble \citep{lin2016scribblesup}, point \citep{bearman2016s}, and image label \citep{papandreou2015emadapt, xu2015learning}. Due to the complete absence of positional information, image-level labels are the cheapest to provide and the most challenging to use, hence this paper will focus on weakly-supervised semantic segmentation (WSSS) from image-level labels. 

Numerous fully-supervised methods have already been proposed and have been reported to perform with impressive accuracy. WSSS researchers consider fully-supervised methods to be the ``upper-bound'' in performance because they are trained with theoretically the most informative supervisory data possible (assuming the annotations are reasonably numerous and accurate) \citep{kwak2017weakly, ye2018learning, kervadec2019constrained}. Indeed, at the time of writing this paper, the best fully-supervised method, DeepLabv3+ \citep{chen2018encoder} attained a 89.0\% mIoU on the PASCAL VOC2012 test set \citep{everingham2010pascal}, which is far higher than the current best weakly-supervised method, IRNet \citep{ahn2019weakly}, with a 64.8\% mIoU. Nonetheless, the quality of WSSS methods is impressive, especially considering that learning to segment without any location-specific supervision is an incredibly difficult task - object extents must be inferred solely from their presence in the training images. Qualitatively, existing WSSS methods deliver excellent segmentation performance on natural scene images while requiring only a fraction of the annotation effort needed for FSS. However, since image labels completely lack positional information, weakly-supervised approaches for natural scene images struggle with three major challenges.

Firstly, WSSS methods struggle to differentiate foreground objects from the background, especially if the background contains strongly co-occurring objects, such as the water from \emph{boat} objects, due to the lack of training information on the precise boundary between them. This was observed by \citep{kolesnikov2016sec, huang2018dsrg, zhou2018weakly} in their qualitative evaluations; \citep{kolesnikov2016improving} addressed the problem by introducing additional model-specific micro-annotations for training. Secondly, WSSS methods can struggle to differentiate frequently co-occurring foreground objects, such as \emph{diningtable} objects from \emph{chair} objects, especially when the scene is cluttered with overlapping objects or the objects consist of components with different appearance; this was observed by \citep{kolesnikov2016sec, zhou2018weakly}. A final challenge is segmenting entire objects instead of discriminative parts, such as the face of a \emph{person} \citep{zhou2018weakly}. Since CNNs tend to identify only discriminative regions for classification, they only generate weak localization cues at those discriminative parts. Using a CNN with a larger field-of-view has been used to alleviate the problem \citep{kolesnikov2016sec}, while others use adversarial erasing \citep{wei2017aepsl} or spatial dropout \citep{lee2019ficklenet} to encourage the CNN to identify less-discriminative regions; still others propagate the localization cues out of discriminative parts using semantic pixel affinities \citep{huang2018dsrg, ahn2019weakly}.

Furthermore, WSSS methods are typically developed solely for natural scene image benchmark datasets, such as PASCAL VOC2012 and little research exists into applying them to other image domains, apart from \citep{yao2016semantic, nivaggioli2019weakly} in satellite images and \citep{xu2014weakly, jia2017constrained} in histopathology images. One might expect WSSS methods to perform similarly after re-training, but these images have many key differences from natural scene images. Natural scene images contain more coarse-grained visual information (i.e. low intra-class variation and high inter-class variation) while satellite and histopathology images contain finer-grained objects (i.e. high intra-class variation and high inter-class variation) \citep{xie2019deep}.  Furthermore, boundaries between objects are often ambiguous and even experts lack consensus when labelling histopathology \citep{xu2017large} and satellite images \citep{mnih2010learning}, unlike in natural scene images. On the other hand, histopathology and satellite images are always imaged at the same scale and viewpoint with minimal occlusion and lighting variations. These differences suggest that WSSS methods cannot be blindly reapplied to different image domains; it is even possible that an entirely different approach to WSSS might perform better in other image domains.

Previously, we proposed a novel WSSS method called HistoSegNet \citep{chan2019histosegnet}, which trains a CNN, extracts weak localization maps, and applies simple modifications to produce accurate segmentation maps on histopathology images. By contrast, WSSS methods developed for natural scene images take the self-supervised learning approach of thresholding the weak localization maps and using them to train a fully-convolutional network. We utilized this approach because the weak localization maps already corresponded well to the entire ground-truth segments in histopathology images, whereas the authors of other WSSS methods attempted self-supervised learning when they observed their weak localization maps corresponding only to discriminative parts in natural scene images. In this paper, we seek to address the lack of research by applying WSSS to different image domains, especially those which are different from natural scene images and share characteristics with histopathology images. This assessment is crucial to determining whether WSSS can be feasibly applied to certain image domains and to discovering the best practices to adopt in difficult image domains. We make the following three main contributions:

\begin{enumerate}
	\item We present a comprehensive review of the literature in multi-class semantic segmentation datasets and weakly-supervised semantic segmentation methods from image labels. For each dataset, we explain the image composition and the annotated classes; for each method, we explain the challenges they attempt to solve and the novel approach that they take.
	\item We implement state-of-the-art WSSS methods developed for natural scene and histopathology images, and then evaluate them on representative natural scene, histopathology, and satellite image datasets. We conduct experiments to compare their quantitative performance and attempt to explain the results by qualitative assessment.
	\item We analyze each approach's compatibility with segmenting different image domains in detail and propose general principles for applying WSSS to different image domains. In particular, we assess: (a) the effect of the sparsity of a classification network's cues, (b) when self-supervised learning is beneficial, and (c) how to address high class co-occurrence in the training data.
\end{enumerate}

The work accomplished in this paper is presented as follows. In Section \ref{sec_relatedwork}, we present a review of the literature in multi-class semantic segmentation datasets and weakly-supervised semantic segmentation methods from image labels. In Section \ref{sec_datasets}, we present the three representative natural scene, histopathology, and satellite image datasets we selected for evaluation; in Section \ref{sec_methods}, we present the state-of-the-art WSSS methods to be evaluated and the modifications we used to ensure fair comparison. In Section \ref{sec_evaluation}, we analyze their performances quantitatively and qualitatively on the selected datasets. In Section \ref{sec_analysis}, we analyze each approach's compatibility with segmenting different image domains in detail and propose general principles for applying WSSS to different image domains. Finally, our conclusions are presented in Section \ref{sec_conclusion}.

\section{Related Work}\label{sec_relatedwork}

\subsection{Multi-class Semantic Segmentation Datasets}

\begin{table*}[h!] 
	\renewcommand{\arraystretch}{1.2} 
	\centering 
	\scriptsize 
	\caption{Multi-Class Semantic Segmentation Datasets, listed in chronological order by image domain. ``Year'' is the year of dataset publication. ``Classes'' is the type of labelled objects under the ``stuff-things'' class distinction (T=Things, S=Stuff, S+T=Stuff and Things). ``\# lbl/img'' is the number of labels per image. ``\# Classes (fg)'' is the total number of possible foreground classes. ``\# Img'' is the total number of original images. ``\# GT'' is the number of images provided with pixel-level annotations. ``Image size'' is the size of the provided original images. ``Resolution'' is the optical resolution of the camera used to capture the original images.}
	\begin{tabular}{lcrccccc} 
		\hlinewd{1.5pt} 
		 Name & Classes & \# lbl/img & \# Classes (fg) & \# Img & \# GT & Image size & Resolution \\ \hlinewd{1.5pt}
		\rowcolor{colour1} \multicolumn{8}{c}{\textbf{Natural Scene Image Datasets}} \\ \hlinewd{1.5pt}
		\rowcolor{colour1} MSRC-21 \citep{shotton2006textonboost} & S+T & $>1$ & 21+void & 591 & 591 & $\sim 320\times 240$ & Variable \\ \hlinewd{.75pt}
		\rowcolor{colour1} SIFT Flow \citep{liu2010sift} & S+T & $>1$ & 30+unlabeled & 2688 & 2688 & $256\times 256$ & Variable \\ \hlinewd{.75pt}		
		\rowcolor{colour1} PASCAL VOC 2012 \citep{everingham2010pascal} & T & $>1$ & 20+bg & 17125 & 10582 & max $=500$ & Variable \\ \hlinewd{.75pt}
		\rowcolor{colour1} PASCAL-Context \citep{mottaghi_cvpr14} & S+T & $>1$ & 59 & 19740 & 10103 & max $\leq 500$ & Variable \\ \hlinewd{.75pt}
		\rowcolor{colour1} COCO 2014 \citep{lin2014microsoft} & T & $>1$ & 80 & 328000 & 123287 & max $\leq 640$ & Variable \\ \hlinewd{.75pt}
		\rowcolor{colour1} ADE20K \citep{zhou2017scene} & S+T & $>1$ & 2693 & 22210 & 22210 & median $640\times 480$ & Variable \\ \hlinewd{.75pt}		
		\rowcolor{colour1} COCO-Stuff \citep{caesar2018coco} & S+T & $>1$ & 172 & 163957 & 163957 & max $\leq 640$ & Variable \\ \hlinewd{1.5pt}
		\rowcolor{colour2} \multicolumn{8}{c}{\textbf{Histopathology Image Datasets}} \\ \hlinewd{1.5pt}
		\rowcolor{colour2} C-Path \citep{beck2011systematic} & S+T & $>1$ & 9+bg & 1286 & 158 & $2256\times 1440$ & $\sim 0.417\upmu$m/px \\ \hlinewd{.75pt}
		\rowcolor{colour2} MMMP (H\&E) \citep{riordan2015automated} & S+T & $>1$ & 17+bg & 102 & 15 & median $2517\times 2434$ & $0.321\upmu$m/px \\ \hlinewd{.75pt}
		\rowcolor{colour2} HMT \citep{kather2016multi} & S+T & $1$ & 7+bg & 5000 & 5000 & $150\times 150$ & $0.495\upmu$m/px \\ \hlinewd{.75pt}
		\rowcolor{colour2} NCT-CRC \citep{kather2019predicting} & S+T & $1$ & 8+bg & 100000 & 100000 & $224\times 224$ & $0.5\upmu$m/px \\ \hlinewd{.75pt}
		\rowcolor{colour2} ADP-morph \citep{hosseini2019atlas, chan2019histosegnet} & S+T & $>1$ & 28+bg & 17668 & 50 & $1088\times 1088$ & $0.25\upmu$m/px \\ \hlinewd{.75pt}
		\rowcolor{colour2} ADP-func \citep{hosseini2019atlas, chan2019histosegnet} & S+T & $>1$ & 4+bg+other & 17668 & 50 & $1088\times 1088$ & $0.25\upmu$m/px \\ \hlinewd{1.5pt}
		\rowcolor{colour3} \multicolumn{8}{c}{\textbf{Visible-light Satellite Image Datasets}} \\ \hlinewd{1.5pt}
		\rowcolor{colour3} UC Merced Land Use \citep{yang2010bag} & S+T & $1$ & 21 & 2100 & 2100 & $256\times 256$ & $1$ ft/px \\ \hlinewd{.75pt}
		\rowcolor{colour3} DeepGlobe Land Cover \citep{demir2018deepglobe} & S & $>1$ & 6+unknown & 1146 & 803 & $2448\times 2448$ & $50$ cm/px \\ \hlinewd{.75pt}
		\rowcolor{colour3} EuroSAT Land Use \citep{helber2019eurosat} & S+T & $1$ & 10 & 27000 & 27000 & $64\times 64$ & $50$ cm/px \\ \hlinewd{1.5pt}
		\rowcolor{colour4} \multicolumn{8}{c}{\textbf{Urban Scene Image Datasets}} \\ \hlinewd{1.5pt}
		\rowcolor{colour4} CamVid \citep{brostow2008segmentation} & S+T & $>1$ & 31+void & 701 & 701 & $960\times 720$ & Fixed \\ \hlinewd{.75pt}
		\rowcolor{colour4} CityScapes \citep{cordts2016cityscapes} & S+T & $>1$ & 30 & 5000 & 3475 & $2048\times 1024$ & Fixed \\ \hlinewd{.75pt}
		\rowcolor{colour4} Mapillary Vistas \citep{neuhold2017mapillary} & S+T & $>1$ & 66 & 25000 & 25000 & $\geq 1920\times 1080$ & Fixed \\ \hlinewd{.75pt}
		\rowcolor{colour4} BDD100K \citep{yu2018bdd100k} & S+T & $>1$ & 40+void & 100000 & 10000 & $1280\times 720$ & Fixed \\ \hlinewd{.75pt}
		\rowcolor{colour4} ApolloScape \citep{wang2019apolloscape} & S+T & $>1$ & 25+unlabeled & 146997 & 146997 & $3384\times 2710$ & Fixed \\ \hlinewd{1.5pt}
	\end{tabular} 
	\label{table_datasets}
\end{table*}

We review below the most prominent multi-class semantic segmentation datasets in four image domains: (1) Natural Scene, (2) Histopathology, (3) Visible-light Satellite, and (4) Urban Scene. Each dataset is listed in Table \ref{table_datasets}; we provide the year of publication, the type of ``stuff-things'' object annotations, the number of labels per image, the number of classes, the total number of images, the number of pixel-level annotated images, the image size, and optical resolution. Further detailed discussion is provided below.

\begin{figure}[h]
	\begin{center}
			 \includegraphics[width=\linewidth]{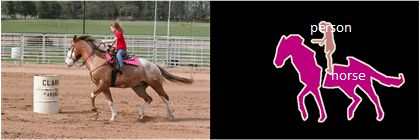}
	\end{center}\vspace{-.1in}
   \caption{Natural Scene Images are captured from natural environments using consumer cameras under vastly varying lighting conditions and viewpoints. The segmentation masks tend to be large and few in number, either leaving large portions of the image unannotated (known as ``things''-only, usually for older datasets) or densely covering the entire image (known as ``stuff and things'', usually for newer datasets) (sample image and ground-truth segmentation from PASCAL VOC2012 \citep{everingham2010pascal}).}
\label{dataset_natural}
\end{figure}

\textbf{Natural Scene Images.}
Natural scene images (also known as ``in the wild'' or ``scene parsing'' images) are captured by consumer cameras under varying light conditions and angles. This terminology is used to emphasize that the images are not synthetically-generated or shot under controlled conditions, as image datasets tended to be in the early days of computer vision research. Occlusion, motion blur, cluttered scenes, ambiguous edges, and multiple scales can be present in these images. MSRC-21 \citep{shotton2006textonboost} is one of the earliest large natural scene datasets annotated at the pixel level, consisting of 591 images (sized $\sim 320\times 240$), each densely annotated with one or more labels selected from 21 object classes (e.g. \emph{building}, \emph{grass}, \emph{tree}), as well as a \emph{void} class. SIFT Flow \citep{liu2010sift} expanded on the number of annotated images and classes; it consists of 2688 images (all sized $256\times 256$), all annotated with 30 foreground classes (and an \emph{unlabeled} class). PASCAL VOC2012 \citep{everingham2010pascal} (see Figure \ref{dataset_natural}) expanded on the number of annotated images even further and subsequently became the benchmark for comparing segmentation algorithms; it consists of 17125 images (with maximum dimension set to 500), 10582 of which are densely annotated with one or more labels selected from 20 foreground classes (e.g. \emph{aeroplane}, \emph{bicycle}, \emph{bird}), as well as a \emph{background} class. The original release provided only 1464 pixel-level annotated set called \emph{train}, but these are typically used with an augmented set to form the 10582 pixel-level annotated set called \emph{trainaug} \citep{bharath2011semantic}.

PASCAL-Context followed up with a dense annotation of the earlier 2010 release of PASCAL VOC, replacing the \emph{background} class with ``stuff'' classes (e.g. \emph{road}, \emph{building}, \emph{sky}); it consists of 19740 images (with maximum dimension $\leq 500$), 10103 of which are labelled with a more manageable subset of 59 labels. COCO 2014 \citep{lin2014microsoft} provided an even larger dataset of ``thing''-annotated images; it consists of 328000 images (with maximum dimension $\leq 640$), 123287 of which are labelled with 80 classes (e.g. \emph{person}, \emph{toilet}, \emph{shoe}), as well as the \emph{background} class. COCO-Stuff (like PASCAL-Context) replaced the \emph{background} class in COCO 2014 with ``stuff'' classes like \emph{grass} and \emph{sky-other}. ADE20K \citep{zhou2017scene} increases the number of classes considered instead of increasing the number of images contained; it consists of 22210 images (median size $640\times 480$), all of which are densely annotated with 2693 classes (e.g. \emph{door}, \emph{table}, \emph{oven}). 

\begin{figure}[h]
	\begin{center}
			 \includegraphics[width=\linewidth]{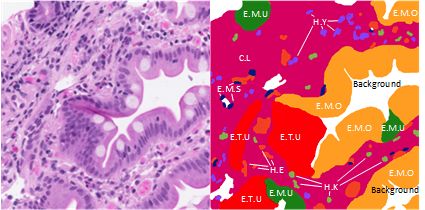}
	\end{center}\vspace{-.1in}
   \caption{Histopathology Images are captured from histological tissue slides scanned with a whole slide imaging scanner and are acquired under strictly controlled lighting conditions and viewpoints. The segmentation masks tend to be small and numerous, densely covering the entire image (known as ``stuff and things'') (sample image and ground-truth segmentation from ADP-morph \citep{hosseini2019atlas}).}
\label{dataset_histopathology}
\end{figure}

\textbf{Histopathology Images.}
Histopathology images are bright-field images of histological tissue slides scanned using a whole slide imaging (WSI) scanner. Although the hematoxylin and eosin (H\&E) stain is most commonly used, staining protocols and scanner types often differ between institutions. The scanned slides are themselves tissue cross sections of three-dimensional specimens stained and preserved inside a glass cover and imaged at the same viewpoint. There is no occlusion (except for folding artifacts) and the background appears uniformly white. Each scanned slide contains vast amounts of visual information, typically to the order of millions of pixels in each dimension. Thus, to reduce the annotation effort, most histopathology datasets are annotated at the patch level rather than the slide level and often each patch is annotated with only one label \citep{kather2016multi, kather2019predicting} or with binary classes \citep{roux2013mitosis, amida2013challenge, monuseg2017challenge, bach2018challenge}. C-Path \citep{beck2011systematic} is likely the first histopathology image datasets to annotate at the pixel-level with multiple classes and multiple labels per image; it consists of 1286 patch images (sized $2256\times 1440$), 158 of which are labelled with at least one of 9 histological types (e.g. \emph{epithelial regular nuclei}, \emph{epithelial cytoplasm}, \emph{stromal matrix}) as well as the \emph{background} class. The H\&E set of MMMP \citep{riordan2015automated} is smaller, but is annotated with more histological types; it consists of 102 images (median size $2517\times 2434$), 15 of which are annotated with one or more of 17 histological types (e.g. \emph{mitotic figure}, \emph{red blood cells}, \emph{tumor-stroma-nuclear}), as well as the \emph{background} class.

HMT \citep{kather2016multi} and NCT-CRC \citep{kather2019predicting} are much larger than C-Path but accomplish this by annotating each image with only one label each. HMT consists of 5000 images (sized $150\times 150$), all labelled with one of 7 histological classes (e.g. \emph{tumour epithelium}, \emph{simple stroma}, \emph{complex stroma}), as well as the \emph{background} class. Ten pixel-level annotated slides (sized $5000\times 5000$) are also provided for evaluation. NCT-CRC consists of 100000 images (sized $224\times 224$), all labelled with one of 8 classes (e.g. \emph{mucus}, \emph{smooth muscle}, \emph{cancer-associated stroma}), as well as the \emph{background} class. ADP \citep{hosseini2019atlas, chan2019histosegnet} (see Figure \ref{dataset_histopathology}) is a histopathology dataset annotated at the pixel level with multiple classes and labels per image; there are 17668 images (sized $1088\times 1088$) in total released with the original dataset \citep{hosseini2019atlas}. All 17668 images are labelled at the image level, and a subset of 50 images is also annotated as a tuning set in a subsequent paper \citep{chan2019histosegnet} with 28 morphological types (known as ``ADP-morph'') and 4 functional types (known as ``ADP-func''). A different subset of 50 images is annotated as an evaluation set and presented in this paper.

\begin{figure}[h]
	\begin{center}
			 \includegraphics[width=\linewidth]{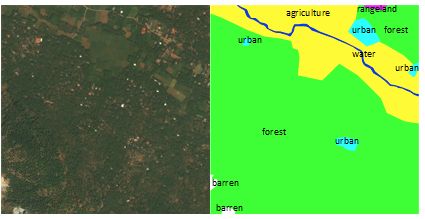}
	\end{center}\vspace{-.1in}
   \caption{Visible-Light Satellite Images are captured from the planetary surface using satellites or airplanes under mildly varying lighting conditions and viewpoints. The segmentation masks tend to be small and numerous, densely covering the entire image (known as ``stuff and things'') (sample image and ground-truth segmentation from DeepGlobe Land Cover \citep{demir2018deepglobe}).}
\label{dataset_satellite}
\end{figure}

\textbf{Visible-Light Satellite Images.}
Visible-light satellite images are images of the Earth taken in the visible-light spectrum by satellites or airplanes. Typically, the surface of the Earth is the object of interest, although occlusion by atmospheric objects (such as clouds) is not uncommon. Lighting conditions can vary, depending on the time of day, and the viewpoint tends not to vary significantly for objects directly below the satellite (distant objects experience distortion due to parallax). Like histopathology images, each satellite image contains vast amounts of visual information, so most satellite image datasets are annotated at the patch level to reduce the annotation cost. UC Merced Land Use \citep{yang2010bag} and EuroSat Land Use \citep{helber2019eurosat} are both annotated with a single label per image. UC Merced Land Use consists of 2100 images (sized $256\times 256$), each labelled with one of 21 land use classes (e.g. \emph{agricultural}, \emph{denseresidential}, \emph{airplane}). EuroSat Land Use, on the other hand, consists of 27000 images (sized $64\times 64$), each labelled with one of 10 land use classes (e.g. \emph{AnnualCrop}, \emph{Industrial}, \emph{Residential}). DeepGlobe Land Cover \citep{demir2018deepglobe} (see Figure \ref{dataset_satellite}) was released for a fully-supervised semantic segmentation challenge and is annotated with multiple labels per image; it comprises of 1146 images (sized $2448\times 2448$), 803 of which are annotated with one or more of 6 classes (e.g. \emph{urban}, \emph{agriculture}, \emph{rangeland}), as well as an \emph{unknown} class.

\begin{figure}[h]
	\begin{center}
			 \includegraphics[width=\linewidth]{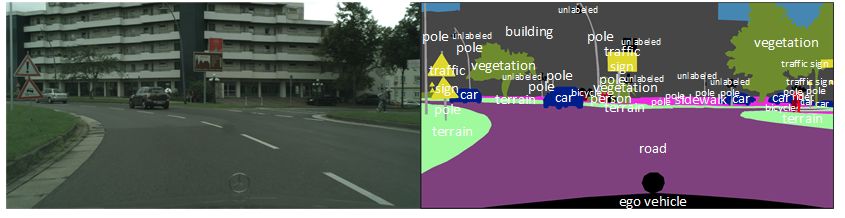}
	\end{center}\vspace{-.1in}
   \caption{Urban Scene images are captured from city environments using a car-mounted camera under vastly varying lighting conditions and viewpoints. The segmentation masks tend to be medium-sized and numerous, densely covering the entire image (known as ``stuff and things'') (sample image and ground-truth segmentation from CityScapes \citep{cordts2016cityscapes}).}
\label{dataset_urban}
\end{figure}

\textbf{Urban Scene Images.}
Urban scene images are images of scenes in front of a driving car, captured by a fixed surveillance camera mounted behind the windshield. Typically, images are captured under different lighting conditions while street-level viewpoint can vary; occlusion is a possibility. The first major urban scene dataset was CamVid \citep{brostow2008segmentation}, which densely annotated all 701 images (sized $960\times 720$) with one or more than labels of 31 urban scene classes (e.g. \emph{Bicyclist}, \emph{Building}, \emph{Tree}), as well as \emph{void} class. CityScapes \citep{cordts2016cityscapes} (see Figure \ref{dataset_urban}) consists of 5000 images (sized $2048\times 1024$), 3475 of which are annotated with 30 classes. Mapillary Vistas \citep{neuhold2017mapillary} is even larger; it consists of 25000 images (sized at least $1920\times 1080$), all annotated with 66 object categories (for semantic segmentation). BDD100K \citep{yu2018bdd100k} consists of a larger set of 100000 images (sized $1280\times 720$), but only 10000 of these are annotated for instance segmentation with 40 object classes (and a \emph{void} class). The April 3, 2018 release of ApolloScape \citep{wang2019apolloscape} is the largest of all to date; it consists of 146997 images (sized $3384\times 2710$), all annotated at the pixel level with 25 classes (and an \emph{unlabeled} class).

\subsection{Weakly-Supervised Semantic Segmentation}
Below, we review the literature in weakly-supervised semantic segmentation from image-level annotations, which refers to learning pixel-level segmentation from image-level labels only. This is the least informative form of weak supervision available for semantic segmentation as it provides no location information for the objects. Different WSSS methods trained with image-level annotations have been proposed to solve this problem; their methodologies can be broadly categorized into four approaches: Expectation-Maximization, Multiple Instance Learning, Object Proposal Class Inference, and Self-Supervised Learning. Table \ref{table_wsss_approach} organizes the reviewed methods by their approaches and common features, while Table \ref{table_wsss_results} lists the methods chronologically with information on the availability of their code online and their segmentation performance in PASCAL VOC 2012, which most of them were developed for.

\begin{table*}[htp]
	\renewcommand{\arraystretch}{1.2}
		\centering
		\scriptsize
		\caption{Weakly-Supervised Semantic Segmentation Methods, organized by approach, with common methodological features and short description for each method.}
		\begin{tabular}{lccccccccl} 
		\hlinewd{1.5pt}
		Method &
		\mcrot{1}{c}{45}{Fully-supervised classification net} &
		\mcrot{1}{c}{45}{Spatial dropout} &
		\mcrot{1}{c}{45}{Expectation-Maximization} &
		\mcrot{1}{c}{45}{CLM inference} &
		\mcrot{1}{c}{45}{Fine contour modification} &
		\mcrot{1}{c}{45}{CLM propagation} &
		\mcrot{1}{c}{45}{Object proposal class inference} &
		\mcrot{1}{c}{45}{Self-supervised segmentation net} &
		\hspace{1in}Method description\\ \hlinewd{1.5pt}
		\rowcolor{colour1} \multicolumn{10}{c}{\textbf{(1) Expectation-Maximization Methods}} \\ \hlinewd{1.5pt}
		\rowcolor{colour1} CCNN \citep{pathak2015constrained} & $\bullet$ &  & $\bullet$ & $\bullet$ & $\bullet$ &  &  & $\bullet$ & Optimize convex latent distribution as pseudo GT; train FCN + CRF \\ \hlinewd{.75pt}
		\rowcolor{colour1} EM-Adapt \citep{papandreou2015emadapt} & $\bullet$ && $\bullet$ & $\bullet$ & $\bullet$ &&& $\bullet$ & Train FCN + predict with class-specific bias to log activation maps + CRF \\ \hlinewd{1.5pt}
		\rowcolor{colour2} \multicolumn{10}{c}{\textbf{(2) Multiple Instance Learning Methods}} \\ \hlinewd{1.5pt}		
		\rowcolor{colour2} MIL-FCN \citep{pathak2014milfcn} & $\bullet$ &&& $\bullet$ &&&&& Train FCN w/ GMP + predict with top prediction at each location, upsample \\ \hlinewd{.75pt}
		\rowcolor{colour2} DCSM \citep{shimoda2016distinct} & $\bullet$ &&& $\bullet$ & $\bullet$ &&&& Train CNN + GBP + depth max + class subtract + multi-scale/layer avg + CRF \\ \hlinewd{.75pt}
		\rowcolor{colour2} BFBP \citep{saleh2016bfbp} & $\bullet$ &&& $\bullet$ & $\bullet$ &&&& Train CNN w/ avg of conv4/5 + fg/bg mask + CRF + LSE  \\ \hlinewd{.75pt}
		\rowcolor{colour2} WILDCAT \citep{durand2017wildcat} & $\bullet$ &&& $\bullet$ & $\bullet$ &&&& Train CNN + class avg of conv feature + pool + local predict + CRF \\ \hlinewd{1.5pt}
		\rowcolor{colour3} \multicolumn{10}{c}{\textbf{(3) Object Proposal Class Inference Methods}} \\ \hlinewd{1.5pt}	
		\rowcolor{colour3} SPN \citep{kwak2017weakly} & $\bullet$ &&& $\bullet$ &&& $\bullet$ & $\bullet$ & Train CNN against GAP and SP as pseudo GT + train FCN \\ \hlinewd{.75pt}
		\rowcolor{colour3} PRM \citep{zhou2018weakly} & $\bullet$ &&& $\bullet$ &&& $\bullet$ && Train CNN w/ PSL + CRM + PB to PRM + predict class for each MCG proposal \\ \hlinewd{1.5pt}
		\rowcolor{colour4} \multicolumn{10}{c}{\textbf{(4) Self-Supervised Learning Methods}} \\ \hlinewd{1.5pt}	
		\rowcolor{colour4} SEC \citep{kolesnikov2016sec} & $\bullet$ &&& $\bullet$ & $\bullet$ &&& $\bullet$ & Train CNN + CAM as pseudo GT + train FCN + predict with CRF \\ \hlinewd{.75pt}
		\rowcolor{colour4} MDC \citep{wei2018revisiting} & $\bullet$ &&& $\bullet$ &&&& $\bullet$ & Train CNN + avg multi-dilated CAM + weigh w/ scores as pseudo GT + train FCN \\ \hlinewd{.75pt}
		\rowcolor{colour4} AE-PSL \citep{wei2017aepsl} & $\bullet$ & $\bullet$ && $\bullet$ &&&& $\bullet$ & Erase DOR during CNN training + CAM as pseudo GT + train FCN \\ \hlinewd{.75pt}
		\rowcolor{colour4} FickleNet \citep{lee2019ficklenet} & $\bullet$ & $\bullet$ && $\bullet$ &&&& $\bullet$ & Train CNN w/ dropout in conv RF + repeat Grad-CAM as pseudo GT + train FCN \\ \hlinewd{.75pt}		
		\rowcolor{colour4} DSRG \citep{huang2018dsrg} & $\bullet$ &&& $\bullet$ & $\bullet$ & $\bullet$ && $\bullet$ & Train CNN + CAM + region growing as pseudo GT + train FCN + predict with CRF \\ \hlinewd{.75pt}
		\rowcolor{colour4} PSA \citep{ahn2018affinitynet} & $\bullet$ &&& $\bullet$ & $\bullet$ & $\bullet$ && $\bullet$ & Train CNN + CAM + random walk in SAG + CRF as pseudo GT + train FCN \\ \hlinewd{.75pt}
		\rowcolor{colour4} IRNet \citep{ahn2019weakly} & $\bullet$ &&& $\bullet$ & $\bullet$ & $\bullet$ && $\bullet$ & Train CNN + CAM + RW in CAM from centroids as pseudo GT + train FCN \\ \hlinewd{1.5pt}				
	\end{tabular}
	\label{table_wsss_approach}
\end{table*}

\begin{table*}[htp] 
	\renewcommand{\arraystretch}{1.2} 
	\centering 
	\scriptsize 
	\caption{Weakly-Supervised Semantic Segmentation Methods (developed for PASCAL VOC2012), by year of publication from 2015 to 2019. Code availability and performance on the PASCAL VOC2012 \emph{val} and \emph{test} sets are also provided for each method.}
	\begin{tabular}{p{4.5cm}p{1.1cm}p{2.5cm}cccc} 
		\hlinewd{1.5pt} 
		 & & & & & \multicolumn{2}{c}{VOC2012 mIoU (\%)} \\
		 Method & Year & Code available? & Train/test code & Code framework & \emph{val} & \emph{test} \\ \hlinewd{1.5pt}
		\rowcolor{colour2} MIL-FCN \citep{pathak2014milfcn} & $2015$ & \href{https://github.com/nightrome/matconvnet-calvin}{Y} & Train/test & MatConvNet & 25.7 & 24.9 \\ \hlinewd{.75pt} 
		\rowcolor{colour1} CCNN \citep{pathak2015constrained} & $2015$ & \href{https://github.com/pathak22/ccnn}{Y} & Train/test & Caffe & 35.3 & 35.6 \\ \hlinewd{.75pt} 
		\rowcolor{colour1} EM-Adapt \citep{papandreou2015emadapt} & $2015$ & Y: \href{https://bitbucket.org/deeplab/deeplab-public/src/master/}{Caffe}, \href{https://github.com/xtudbxk/em-adapt-tensorflow}{TensorFlow} & Train/test & Caffe, TensorFlow & \textbf{38.2} & \textbf{39.6} \\ \hlinewd{.75pt} 
		\rowcolor{colour2} DCSM w/o CRF \citep{shimoda2016distinct} & $2016$ & \href{https://github.com/shimoda-uec/dcsm}{Y} & Test & Caffe & 40.5 & 41 \\ \hlinewd{.75pt} 
		\rowcolor{colour2} DCSM w/ CRF \citep{shimoda2016distinct} & $2016$ & \href{https://github.com/shimoda-uec/dcsm}{Y} & Test & Caffe & 44.1 & 45.1 \\ \hlinewd{.75pt} 
		\rowcolor{colour2} BFBP \citep{saleh2016bfbp} & $2016$ & N & No & - & 46.6 & 48.0 \\ \hlinewd{.75pt} 
		\rowcolor{colour4} SEC \citep{kolesnikov2016sec} & $2016$ & Y: \href{https://github.com/kolesman/SEC}{Caffe}, \href{https://github.com/xtudbxk/SEC-tensorflow}{TensorFlow} & Train/test & Caffe, TensorFlow & \textbf{50.7} & \textbf{51.7} \\ \hlinewd{.75pt} 
		\rowcolor{colour2} WILDCAT + CRF \citep{durand2017wildcat} & $2017$ & \href{https://github.com/durandtibo/wildcat.pytorch}{Y} & Train/test & PyTorch & 43.7 & - \\ \hlinewd{.75pt} 
		\rowcolor{colour3} SPN \citep{kwak2017weakly} & $2017$ & \href{https://github.com/parag2489/keras_superpixel_pooling}{Y} & Custom layer only & Keras & 50.2 & 46.9 \\ \hlinewd{.75pt} 
		\rowcolor{colour4} AE-PSL \citep{wei2017aepsl} & $2017$ & N & No & - & \textbf{55.0} & \textbf{55.7} \\ \hlinewd{.75pt} 
		\rowcolor{colour3} PRM \citep{zhou2018weakly} & $2018$ & \href{https://github.com/ZhouYanzhao/PRM}{Y} & Test & PyTorch & 53.4 & - \\ \hlinewd{.75pt} 
		\rowcolor{colour4} DSRG (VGG16) \citep{huang2018dsrg} & $2018$ & Y: \href{https://github.com/speedinghzl/DSRG}{Caffe}, \href{https://github.com/xtudbxk/DSRG-tensorflow}{TensorFlow} & Train/test & Caffe, TensorFlow & 59.0 & 60.4 \\ \hlinewd{.75pt} 
		\rowcolor{colour4} PSA (DeepLab) \citep{ahn2018affinitynet} & $2018$ & \href{https://github.com/jiwoon-ahn/psa}{Y} & Train/test & PyTorch & 58.4 & 60.5 \\ \hlinewd{.75pt} 
		\rowcolor{colour4} MDC \citep{wei2018revisiting} & $2018$ & N & No & - & 60.4 & 60.8 \\ \hlinewd{.75pt} 
		\rowcolor{colour4} DSRG (ResNet101) \citep{huang2018dsrg} & $2018$ & Y: \href{https://github.com/speedinghzl/DSRG}{Caffe}, \href{https://github.com/xtudbxk/DSRG-tensorflow}{TensorFlow} & Train/test & Caffe, TensorFlow & 61.4 & 63.2 \\ \hlinewd{.75pt} 
		\rowcolor{colour4} PSA (ResNet38) \citep{ahn2018affinitynet} & $2018$ & \href{https://github.com/jiwoon-ahn/psa}{Y} & Train/test & PyTorch & \textbf{61.7} & \textbf{63.7} \\ \hlinewd{.75pt} 
		\rowcolor{colour4} FickleNet \citep{lee2019ficklenet} & $2019$ & N & No & - & 61.2 & 61.9 \\ \hlinewd{.75pt} 
		\rowcolor{colour4} IRNet \citep{ahn2019weakly} & $2019$ & \href{https://github.com/jiwoon-ahn/irn}{Y} & Train/test & PyTorch & \textbf{63.5} & \textbf{64.8} \\ \hlinewd{1.5pt} 
	\end{tabular}
	\label{table_wsss_results}
\end{table*}

\textbf{(1) Expectation-Maximization.}
The Expectation - Maximization approach consists of alternately optimizing a latent label distribution across the image and learning a segmentation of the image from that latent distribution. In practice, this means starting with a prior assumption about the class distribution (e.g. the size of each class segment) from the ground-truth image annotations, training a Fully Convolutional Network (FCN) to replicate these inferred segments, updating the prior assumption model based on the FCN features, and repeating the training cycle again.

CCNN \citep{pathak2015constrained} uses block coordinate descent to alternate between (1) optimizing the convex latent distribution of fixed FCN outputs with segment-specific constraints (e.g. for suppressing absent labels and encouraging large foreground segments) and (2) training a FCN with SGD against the fixed latent distribution. EM-Adapt \citep{papandreou2015emadapt} alternates between (1) training a FCN with class-specific bias to each activation map with global sum pooling on the log activation maps to train against the image-level labels and (2) adaptively setting the class biases to equal a fixed percentile of the score difference between the maximum and class score at each position (in order to place a lower bound on the segment area of each class).

\textbf{(2) Multiple Instance Learning.}
The Multiple Instance Learning (MIL, or Bag of Words) approach consists of learning to predict the classes present in an image (known as a ``bag'') given ground-truth image-level annotations and then, given the knowledge that at least one pixel of each class is present, assigning pixels (known as ``words'') to each predicted class. In practice, this often means training a Convolutional Neural Network (CNN) with image-level loss and inferring the image locations responsible for each class prediction.

MIL-FCN \citep{pathak2014milfcn} trains a FCN headed by a $1\times 1$ conv layer and a Global Max Pooling (GMP) layer against the image-level annotations, then at test time, it predicts the top class at each location in the convolutional features and the predicted class map is bilinearly upsampled. DCSM \citep{shimoda2016distinct} trains a CNN at the image level and uses GBP (guided back-propagation) to obtain the coarse class activation maps at the upper intermediate convolutional layers, then subtracts the maps from each other, and takes the average of the maps across different scales and layers, followed by CRF post-processing. BFBP \citep{saleh2016bfbp} trains a FCN with a foreground/background mask generated by CRF on the scaled average of conv4 and conv5 features with cross-entropy loss between the image-level annotations and the LSE pool of foreground- and background-masked features; CRF post-processing is applied at test time. WILDCAT \citep{durand2017wildcat} trains a FCN with conv5 features being fed into a WSL transfer network, then applies class-wise average pooling and weighted spatial average of top- and lowest-activating activations; at test time, it infers the maximum-scoring class per position and post-processes with CRF.

\textbf{(3) Object Proposal Class Inference.}
The Object Proposal Class Inference approach often takes elements from both the MIL and Self-Supervised Learning approaches but starts by extracting low-level object proposals and then assigns the most probable class to each one using coarse-resolution class activation maps inferred from the ground-truth image-level annotations. SPN \citep{kwak2017weakly} trains a CNN which performs a spatial average of the features closest to each superpixel from the original image and then has FC classifier layers with an image-level loss, and these superpixel-pooled features are then used as pseudo ground-truths to train a FCN. PRM \citep{zhou2018weakly} extracts MCG (Multi-scale Combinatorial Grouping) low-level object proposals, trains a FCN with peak stimulation loss, then peak backpropagation is done for each peak in the Class Response Map to obtain the Peak Response Map. Each object proposal is then scored using the PRM peaks and assigned the top-ranked classes with non-maximum suppression.

\textbf{(4) Self-Supervised Learning.}
The Self-Supervised Learning approach is similar to the MIL approach but uses the inferred pixel-level activations as pseudo ground-truth cues (or seeds) for self-supervised learning of the final pixel-level segmentation maps. In practice, this usually means training a ``backbone'' classification network to produce Class Activation Map (CAM) seeds and then training a FCN segmentation network on these seeds. SEC \citep{kolesnikov2016sec} is the prototypical method to take this approach; it trains a CNN and applies CAM to produce pseudo ground-truth segments to train a FCN against the generated seeds, against the image-level label, and a constraint loss against the CRF-processed maps. MDC \citep{wei2018revisiting} takes a similar but more multi-scale approach by training a CNN with multi-dilated convolutional layers at the image level, adding multi-dilated block CAMs together, and then generating pseudo ground-truths to train a FCN with the class score-weighted maps. However, methods taking this approach tend to produce good segmentations only for discriminative parts rather than entire objects, so different solutions have been suggested to fill the low-confidence regions in between.

One solution is to apply adversarial or stochastic erasing during training and encourage the networks to learn less discriminative object parts. AE-PSL \citep{wei2017aepsl} generates CAMs as pseudo ground-truths for training a FCN just like SEC, but during CNN training, high-activation regions from the CAMs are adversarially erased from the training image. FickleNet \citep{lee2019ficklenet}, on the other hand, trains a CNN at the image level with centre-fixed spatial dropout in the later convolutional layers (by dropping out non-centre pixels in each convolutional window) and then runs Grad-CAM multiple times to generate a thresholded pseudo ground-truth for training a FCN.

Another solution is to simply propagate class activations from high-confidence regions to adjacent regions with similar visual appearance. DSRG \citep{huang2018dsrg} trains a CNN and applies region-growing on the generated CAMs to produce a pseudo ground-truth for training a FCN. PSA \citep{ahn2018affinitynet} similarly trains a CNN but propagates the class activations by performing a random walk from the seeds in a semantic affinity graph as a pseudo ground-truth for training a FCN. IRNet \citep{ahn2019weakly} is similar as well, but seeks to segment individual instances by performing the random walk from low-displacement field centroids in the CAM seeds up until the class boundaries as the pseudo ground-truths for training a FCN. It is significant to note that, judging from their quantitative performance on PASCAL VOC2012-\emph{val}, the top five performing WSSS methods all use the self-supervised learning approach, and three of these additionally use the outward class propagation technique.

\subsection{Semantic Segmentation Methods for Satellite and Histopathology Images}

\textbf{Satellite Images.}
Compared to natural scene images, relatively limited research has been conducted in multi-class semantic segmentation in satellite images. Most work has been done with fully-supervised learning, since these annotations are the most informative. Indeed, the best performing methods tend to use variants of popular methods developed for natural scene images. In the DeepGlobe Land Cover Classification challenge \citep{demir2018deepglobe}, for instance, DFCNet \citep{tian2018dense} is the best performing method and is a variant of the standard FCN \citep{long2015fully} with multi-scale dense fusion blocks and auxiliary training on the road segmentation dataset. The second-best method Deep Aggregation Net \citep{kuo2018deep} is DeepLabv3 \citep{chen2017rethinking} with Gaussian filtering applied to the segmentation masks and graph-based post-processing to remove small segments (by assigning them to the class of their top-left neighbouring segment if their size falls below a threshold). The third-best method \citep{seferbekov2018feature} uses a variant of FPN \citep{lin2017feature}, but the convolutional branch networks attached to the intermediate convolutional layers (known as RPN heads in the original FPN method for proposing object regions) with skip connections are instead used to output multi-scale features that are concatenated into a final segmentation map (at the original image resolution). Another assessment of different semantic segmentation techniques on the even larger NAIP dataset used the standard DenseNet and U-Net architectures without significant modifications \citep{robinson2019large}. For weakly-supervised learning, even less research is published; what research can be found attempts to apply standard WSSS techniques to satellite images. Indeed, the state-of-the-art Affinity-Net (or PSA) was adapted by \citep{nivaggioli2019weakly} for segmenting DeepGlobe images with only image-level annotations (while experimenting with de-emphasizing background loss and eliminating the background class altogether). SDSAE \citep{yao2016semantic}was used to train on image-level land cover annotations on the LULC set as auxiliary data and the trained parameters were then transferred to perform pixel-level segmentation on their proposed Google Earth land cover dataset.

\textbf{Histopathology Images.}
In histopathological images, semantic segmentation methods tend to address binary-class problems, probably due to the significant expense of annotating large histopathology images with multiple classes. These tend to label each pixel with either diagnoses (e.g. cancer/non-cancer \citep{bach2018challenge}) tissue/cell types (e.g. gland \citep{sirinukunwattana2017gland}, nuclei \citep{monuseg2017challenge}, and mitotic/non-mitotic figures \citep{roux2013mitosis, amida2013challenge}). As with satellite imagery, semantic segmentation methods for histopathology tend to use fully-supervised learning. Sliding patch-based methods have been used to segment mitotic figures \citep{ciresan2013mitosis, malon2013classification}, cells \citep{shkolyar2015automatic}, neuronal membranes \citep{ciresan2012neuronal}, and glands \citep{li2016gland, kainz2015colon}. Superpixel-based object proposal methods have been used to segment tissues by histological type \citep{xu2016dcnn, turkki2016til}. Fully convolutional methods have been used by training a FCN with optional contour post-processing \citep{chen2016dcan, lin2017scannet}. Weakly-supervised methods, on the other hand, are much rarer and tend to use a patch-based MIL approach. MCIL was developed to segment colon TMAs by cancer grade with only image-level annotations by clustering the sliding patch features \citep{xu2014weakly}. EM-CNN \citep{hou2016patch} is trained on slide-level cancer grade annotations and predicts at the patch level and forms a decision fusion model afterward to predict the cancer grade of the overall slide. Although the pre-decision segmentation map only has patch-level resolution, it could theoretically be extended to pixel-level resolution had the patches been extracted densely at test time. DWS-MIL \citep{jia2017constrained} trains a binary-class CNN with multi-scale loss against the image-level labels by assuming the same label throughout each ground-truth image (essentially using Global Average Pooling (GAP)). ScanNet \citep{lin2018scannet} is a FCN variant trained on patch-level prediction; at test time, a block of multiple patches is inputted to the network and a coarse pixel-level segmentation is outputted; originally developed for breast cancer staging, it has also been applied to lung cancer classification \citep{wang2018weaklysupervised}. HistoSegNet \citep{chan2019histosegnet} trains a CNN on patch-level histological type annotations and applies Grad-CAM to infer coarse class maps, followed by class-specific modifications (\emph{background} and \emph{other} class map augmentation, class map subtraction), and post-processing with CRF to produce fine pixel-level segmentation maps.

\section{Datasets}\label{sec_datasets}

At the time of writing, the vast majority of WSSS algorithms have been developed for natural scene images. Hence, to analyze their performance on other image domains, we selected three representative datasets for evaluation: (1) Atlas of Digital Pathology (histopathology), (2) PASCAL VOC2012 (natural scene), and (3) DeepGlobe Land Cover Classification (satellite).


\subsection{Atlas of Digital Pathology (ADP)}

The Atlas of Digital Pathology \citep{hosseini2019atlas} is a database of histopathology patch images (sized $1088\times 1088$) extracted from WSI scans of healthy tissues stained by the same institution and scanned from different organs with the Huron TissueScope LE1.2 scanner ($0.25\mu$m/pixel resolution). This dataset was selected due to the large quantity of image-labelled histopathology patches available for training, each labelled with 28 morphological types (with \emph{background} added for segmentation) and 4 functional types (with \emph{background} and \emph{other} added for segmentation). We use the \emph{train} set of 14,134 image-annotated patches for training; for validation, we use the \emph{tuning} set of 50 pixel-annotated patches, which has more classes per image and was to tune HistoSegNet \citep{chan2019histosegnet}; for evaluation, we use the \emph{segtest} set of 50 pixel-annotated patches.

\subsection{PASCAL VOC2012}

The 2012 release of the PASCAL VOC challenge dataset \citep{everingham2010pascal} consists of natural scene (``in the wild'') images captured by a variety of consumer cameras. This dataset was selected due to its status as the default benchmark set for WSSS algorithms. Each image is labelled with 20 foreground classes, with an added \emph{background} class for segmentation. For training, we use the \emph{trainaug} set of 12,031 image-annotated images \citep{bharath2011semantic}; for evaluation, we use the \emph{val} set of 1,449 pixel-annotated images (the segmentation challenge ranks methods with the \emph{test} set of 1,456 un-annotated images through the evaluation server).

\subsection{DeepGlobe Land Cover Classification}

The DeepGlobe Land Cover Classification dataset consists of visible-light satellite images extracted from the Digital-Globe+Vivid Images dataset \citep{demir2018deepglobe}. This dataset was selected due to its status as the only multi-label satellite dataset for segmentation. Each image is labelled with 6 land cover classes (and an \emph{unknown} class for non-land cover regions). For training, we randomly split the \emph{train} set of 803 pixel-annotated images into our own 75\% training set of 603 image-annotated images and 25\% test set of 200 pixel-annotated images. The \emph{unknown} class was omitted for both training and evaluation.

\section{Methods}\label{sec_methods}

\begin{figure*}[htp]
	\begin{center}
			 \includegraphics[width=.95\textwidth]{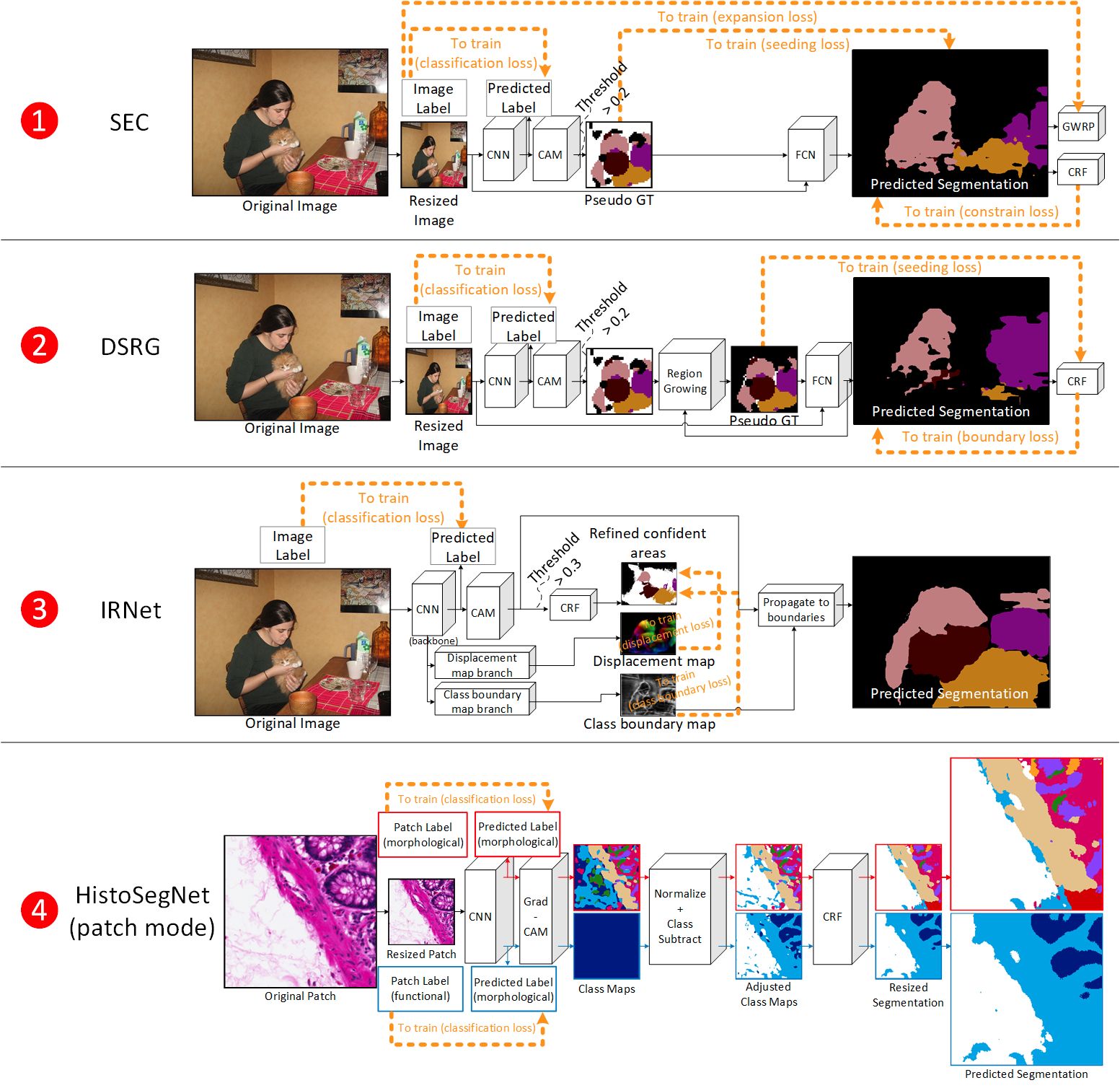}
	\end{center}\vspace{-.1in}
   \caption{Overview of the four compared WSSS methods: (1) SEC, (2) DSRG, (3) IRNet, and (4) HistoSegNet. All four methods first train a classification CNN on the image labels, produce coarse activation maps, and then process these maps to infer fine-grained pixel labels. SEC, DSRG, and IRNet are self-supervised learning methods developed for PASCAL VOC2012; they produce fine-grained segmentations by training a downstream neural network model with the coarse maps as pseudo ground-truths. HistoSegNet is an object proposal class inference method developed for ADP; it produces fine-grained segmentation by applying a hand-tuned dense conditional random field to the coarse maps.}
\label{fig_methods}
\end{figure*}

To compare WSSS algorithm performance on the selected datasets, three state-of-the-art methods were chosen: (1) SEC, (2) DSRG, and (3) HistoSegNet. SEC and DSRG were both developed for natural scene images (PASCAL VOC2012) and had both the highest mean Intersection-over-Union (mIoU) at the time of writing and had code implementations available online; HistoSegNet was developed for histopathology images (ADP) and is the only WSSS method developed specifically for non-natural scene images. Furthermore, SEC and DSRG share a common self-supervised FCN training approach while HistoSegNet uses a simpler Grad-CAM refinement approach. See \ref{fig_methods} for an overview of the three evaluated methods.


\subsection{Seed, Expand and Constrain (SEC)}

Seed, Expand and Constrain (SEC) \citep{kolesnikov2016sec} was developed for the PASCAL VOC2012 dataset and consists of four trainable stages: (1) a classification CNN is trained on image labels, (2) CAMs are generated from the trained CNN, (3) the CAMs are thresholded and overlap conflicts resolved as seeds/cues, and (4) the seeds are used for self-supervised training of a FCN (DeepLabv1, also known as DeepLab-LargeFOV \citep{chen2014semantic}).

\textbf{(1) Classification CNN.} First, two classification CNNs are trained on the annotated images: (1) the ``foreground'' network (a variant of the VGG16 network omitting the last two pooling layers and the last two fully-connected layers and replacing the flattening layer with a GAP layer) and (2) the ``background'' network (a variant of the VGG16 network omitting the last two convolutional blocks).

\textbf{(2) CAM.} The Class Activation Map (CAM) is then applied to both the ``foreground'' and ``background'' networks for each image in the \emph{trainaug} dataset.

\textbf{(3) Seed Generation.} For the ``foreground'' network, each class CAM is thresholded above 20\% of the maximum activation as a weak localization cue (or seed); for ``background'' network, the class CAMs are added, a 2D median filter is applied, and the 10\% lowest-activating pixels are thresholded as the additional \emph{background} cue. In regions where cues overlap, the class with the smaller cue takes precedence.

\textbf{(4) Self-Supervised FCN Learning.} Finally, these weak localization cues are used as pseudo ground-truths for self-supervised learning of a Fully Convolutional Network (FCN) \citep{long2015fully}. A three-part loss function is used on the FCN output: (1) a seeding loss with the weak cues, (2) an expansion loss with the image labels, and (3) a constrain loss with itself after applying dense CRF. At test time, dense CRF is used for post-processing.

\subsection{Deep Seeded Region Growing (DSRG)}

Deep Seeded Region Growing (DSRG) \citep{huang2018dsrg} was, similarly to SEC, also developed for PASCAL VOC2012 and takes the similar approach of generating weak seeds using CAM for training a FCN (this time, DeepLabv2, also known as DeepLab-ASPP \citep{chen2017deeplab}). However, this method differs in several important ways. First, there is no ``background'' network - the \emph{background} activation is instead generated separately using the fixed DRFI method \citep{jiang2013salient}. Secondly, the foreground CAMs are thresholded above 20\% of the maximum activation and then used as seeds for convolutional feature-based region growing into a weak localization cue. Thirdly, a two-part loss function is used on the FCN output: (1) a seeding loss with the region-grown weak cues and (2) a boundary loss with itself after applying dense CRF (identical to constrain loss in SEC). Again, dense CRF is applied at test time.


\subsection{IRNet}

Inter-pixel Relation Network (IRNet) \citep{ahn2019weakly} was developed for both semantic and instance segmentation in PASCAL VOC2012, although we only consider the semantic segmentation case. While it utilizes CAMs as pseudo ground-truths like SEC and DSRG, it trains two branches from the backbone network to predict auxiliary information instead of the pixel classes directly. The method consists of the following five stages:

\textbf{(1) Classification CNN \& (2) CAM.} Similarly to SEC and DSRG, a classification CNN is first trained on the labelled images (ResNet50 architecture \citep{he2016deep}) and CAMs are generated after training is complete.

\textbf{(3) Seed Generation.} The CAM of each confident class is then thresholded above $0.3$ and refined with dense CRF as foreground seeds. Regions with CAM confidence below $0.05$ and left without a foreground seed after refining with dense CRF are considered as background seeds.

\textbf{(4) Self-Supervised DF and CBM Learning.} The foreground and background seeds are used as pseudo ground-truths for training two branches from the backbone network: (1) a displacement field (DF) to predict the positional displacement of each pixel from each seed instance's centroid and (2) a class boundary map (CBM) to predict the likelihood of a class boundary existing at each pixel, by maximizing the value between neighbouring pixels (within a set radius) seeded with different classes and minimizing it for pixels of the same seed class.

\textbf{(5) CAM Random Walk Propagation with CBM.} Finally, the CAM of each confident class is propagated by random walk with the inverse of the class boundary map as the transition probability matrix. This enables confident CAM regions to propagate into less confident regions inside likely class boundaries.

\subsection{HistoSegNet}


The HistoSegNet algorithm \citep{chan2019histosegnet} was developed for the ADP database of histological tissue type (HTT), and consists of four stages: (1) a classification CNN is trained on patch-level annotations, followed by (2) a hand-crafted Grad-CAM, (3) activation map adjustments (e.g. \emph{background} / \emph{other} activations, class subtraction), and (4) a dense CRF. By default, HistoSegNet accepts $224\times 224$-pixel patches that are resized from a scan resolution of $0.25\times\frac{224}{1088}=1.2143\mu$m/pixel. Processing is conducted mostly independently for the morphological and functional segmentation modes. Patch predictions between stages (3) and (4) to minimize boundary artifacts.

\textbf{(1) Classification CNN.} First, a classification CNN is trained on the HTT-labelled patches of the ADP database (i.e. the $31$ HTTs in the third level, excluding undifferentiated and absent types). The architecture is a variant of VGG-16, except: (1) the softmax layer is replaced by a sigmoid layer, (2) batch normalization is added after each convolutional layer activation, and (3) the flattening layer is replaced by a global max pooling layer. Furthermore, no color normalization was applied since the same WSI scanner and staining protocol were used for all images.

\textbf{(2) Grad-CAM.} To infer pixel-level HTT predictions from the pre-trained CNN, Gradient-Weighted Class Activation Maps (Grad-CAM) \citep{selvaraju2017grad} are applied; this is a generalization of Class Activation Map (CAM) \citep{zhou2016learning} for all CNN architectures. Grad-CAM scores each pixel in the original image by its importance for a CNN's class prediction. The Grad-CAM provides coarse pixel-level class activation maps for each image which are scaled from 0 to 1 and multiplied by their HTT confidence scores for stability.

\textbf{(3) Inter-HTT Adjustments.} The original ADP database has no non-tissue labels, so \emph{background} maps must be produced for both morphological and functional modes; ADP also omits non-functional labels for the functional mode, so \emph{other} maps must also be produced. This allows HistoSegNet to avoid making predictions where no valid pixel class from ADP exists. The \emph{background} activation is assumed to be regions of high white illumination which are not transparent-staining tissues (e.g. white/brown adipose, glandular/transport vessels); it is generated by applying a scaled-and-shifted sigmoid to the mean-RGB image, then subtracting the transparent-staining class activations, and applying a 2D Gaussian blur. The \emph{other} activation is assumed to be regions of low activation for the \emph{background} and all other functional tissues; it is generated by taking the 2D maximum of: (1) all other functional type activations, (2) white and brown adipose activations (from the morphological mode), and (3) the background activation. Then, this probability map is subtracted from one and scaled by 0.05. Finally, overlapping Grad-CAMs are differentiated by subtracting each activation map from the 2D maximum of the other Grad-CAMs - in locations of overlap, this suppresses weak activations overlapping with strong activations and improves results for dense CRF.

\textbf{(4) Dense CRF.} The resultant activation maps are still coarse and poorly conform to object contours, so the dense Conditional Random Field (CRF) \citep{krahenbuhl2011efficient} is used, with an appearance kernel and a smoothness kernel being applied for $5$ iterations using different settings each for the morphological and functional modes.

\begin{figure}[h!]
	\begin{center}
			 \includegraphics[width=\linewidth]{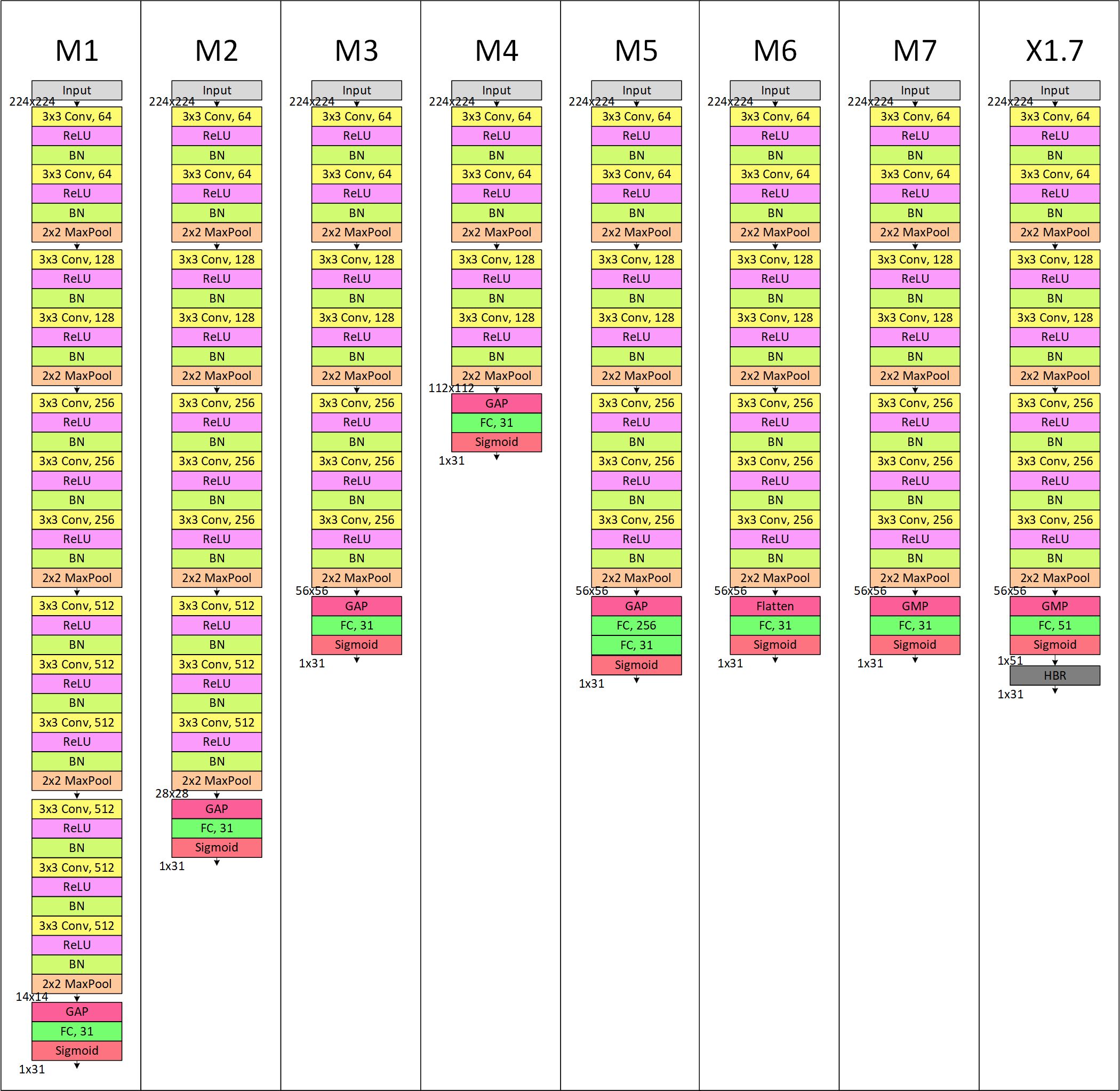}
	\end{center}
   \caption{Overview of the eight ablative architectures used to study the effect of network architecture on WSSS performance.}
\label{fig_model_architectures}
\end{figure}

\textbf{Ablative Study.} SEC and DSRG use VGG16 to generate weak localization seeds while HistoSegNet uses a much shallower 3-block VGG16 variant; this raises the question, ``Is network architecture important for WSSS performance?'' This issue has never been explored before, so to answer this, we analyzed the performance of HistoSegNet using eight variant architectures of VGG16, named M1 (i.e. VGG16), M2, M3, M4, M5, M6, M7, and X1.7 (i.e. the one used in HistoSegNet) (see Figure \ref{fig_model_architectures}). M1 through M4 analyze the effect of \emph{network depth}: they all use GAP for vectorization and a single fully-connected layer, but differ in the number of convolutional blocks: 5, 4, 3, and 2 respectively. M5 through M7, on the other hand, analyze the effect of the \emph{vectorization operation}: they all have 3 convolutional blocks and a single fully-connected layer, but use GAP, Flatten, and GMP for vectorization respectively. Finally, X1.7 analyzes the effect of \emph{hierarchical binary relevance} (HBR) \citep{tsoumakas2009mining}: it is identical to M7 but trains on all 51 classes of the ADP class set and tests on only the 31 segmentation classes. All eight networks were trained on Keras (TensorFlow backend) for 80 epochs with cyclical learning rate and a batch size of 16; they were evaluated for classification on the test set and for segmentation on the \emph{segtest} set (both ADP-morph and ADP-func).

\begin{figure}[h!]
	\centerline{
		\subfigure[Classification performance]{\includegraphics[width=\linewidth]{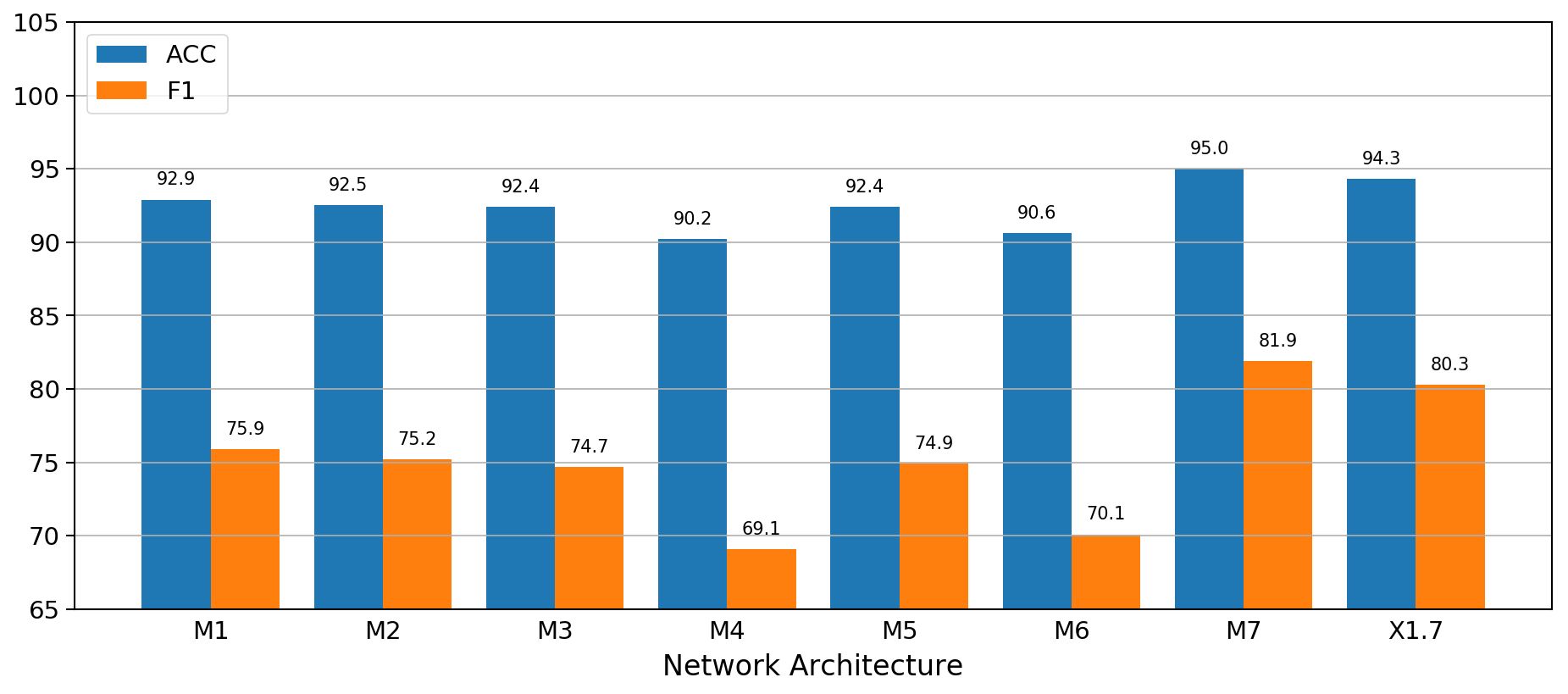}\label{fig_model_class}}
	}
	\centerline{
		\subfigure[Segmentation performance (ADP-morph)]{\includegraphics[width=\linewidth]{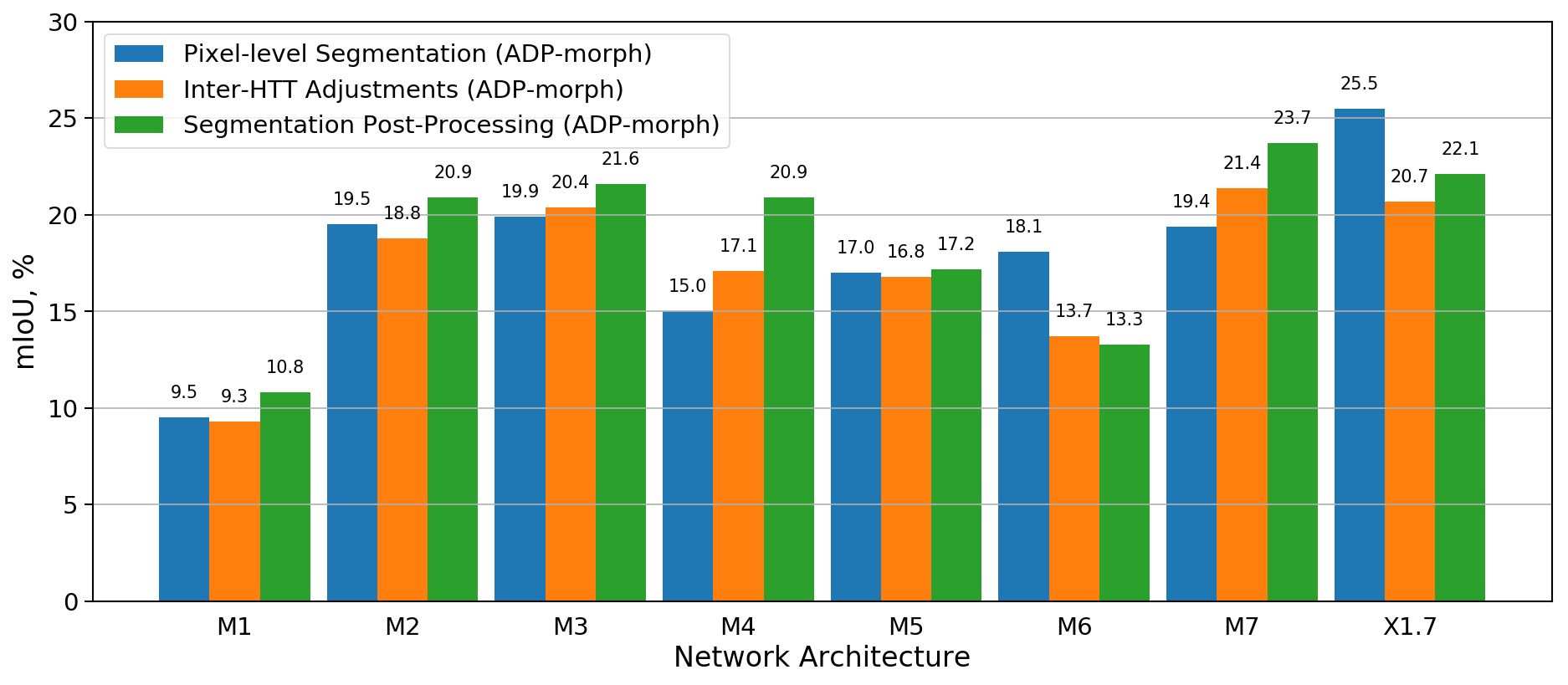}\label{fig_model_seg_morph}}
	}
	\centerline{
		\subfigure[Segmentation performance (ADP-func)]{\includegraphics[width=\linewidth]{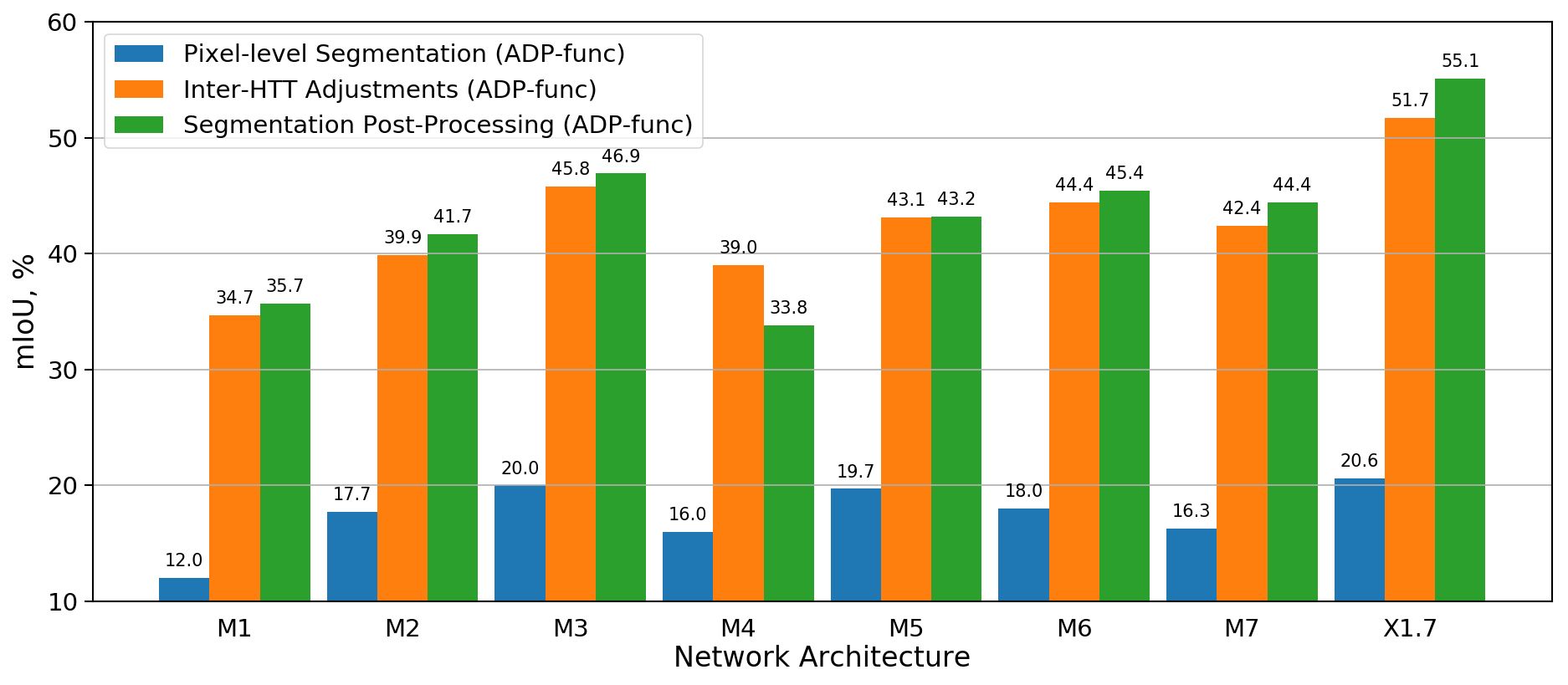}\label{fig_model_seg_func}}
	}
   \caption{Performance of the eight ablative architectures in (a) classification, (b) morphological segmentation, and (c) functional segmentation. Deeper networks (i.e. M1) perform better than shallow networks (i.e. M2-M4) and vectorizing with GMP (i.e. M7) is better than GAP or Flatten (i.e. M5, M6) for classification, but shallower networks are better for segmentation up to 3 blocks (i.e. M3). Note that some scales do not start at zero.}
\label{fig_model_performances}
\end{figure}

For classification (Figure \ref{fig_model_class}), networks with greater depth (i.e. M1) predict better than those with lesser depth (i.e. M2-M4), networks vectorized with GMP (i.e. M7) predict better than with GAP (i.e. M5) and Flatten (i.e. M6), and networks without HBR (i.e. M7) predict better than those with HBR (i.e. X1.7). But for segmentation, a different pattern emerges. For the morphological types (Figure \ref{fig_model_seg_morph}), although GMP vectorization and no HBR (i.e. M7) are still superior, lesser depth is beneficial up to 3 blocks (i.e. M3). For the functional types (Figure \ref{fig_model_seg_func}), lesser depth is also beneficial up to 3 blocks (i.e. M3), but Flatten vectorization (i.e. M6) and with HBR (i.e. X1.7) are superior in this case. These results show that the classification network design is important for subsequent WSSS performance and that deeper networks such as VGG16 may perform well on classification but fail on segmentation due to their smaller convolutional feature maps.

\section{Performance Evaluation}\label{sec_evaluation}

In this section, the four state-of-the-art methods are modified for the three representative segmentation datasets and their relative performance is evaluated. Until this point, there have been few attempts to apply WSSS methods to different image domains: SEC, DSRG, and IRNet have been developed for PASCAL VOC2012, while HistoSegNet has been developed for ADP. Hence, it is imperative to assess whether certain methods out-perform others on different segmentation datasets.

\subsection{Setup}
The original WSSS methods were developed to use different classification CNN architectures: VGG16 for SEC and DSRG, the shallower X1.7 for HistoSegNet, and the deeper ResNet-50 for IRNet. To avoid the possibility that the classification CNN choice would unfairly favour certain methods over others, we chose to implement all four WSSS methods with each of VGG16 and X1.7 - eight network-method configurations result. As Hierarchical Binary Relevance (HBR) was used in X1.7 to leverage the hierarchical class taxonomy in ADP, it is omitted for the non-hierarchical datasets (PASCAL VOC2012 and DeepGlobe) and denoted as M7. To generate seeds for the self-supervised methods, we decided to select confidence thresholds for SEC and DSRG that ensured less than $50\%$ of the training set images were covered by seeds, which was heuristically determined to be optimal (this will be covered in more detail in Section \ref{sec_a2}). For IRNet, confidence thresholds were tuned using coordinate descent.

In common with the original practice used in SEC, DSRG, and HistoSegNet, images were first resized to $321\times 321$ and $224\times 224$ for VGG16 and M7 respectively, with flipping and moderate scaling used as image augmentation. We neither explored other image re-sizing techniques nor different receptive fields for the sake of simplicity. All CNNs were trained to convergence in 80 epochs with cyclical learning rate \citep{smith2017cyclical} (triangular policy, between 0.001 and 0.02 with a period of 10 epochs	and 0.5 decay every 20 epochs). For SEC and DSRG, a simple stepwise decaying learning rate was used, starting at 0.0001 and 0.5 decay every 4 epochs. A constant learning rate of 0.1 and weight decay of 0.0001 was used for training IRNet for 3 epochs, following the authors' settings. All trainable weights in the CNNs, SEC, DSRG, and IRNet were pre-initialized initialized from ImageNet; this improved performance, even for ADP and DeepGlobe. See Section 2 of the Supplementary Materials for the CNN training details, Section 5 for SEC and DSRG.

Furthermore, non-foreground objects (e.g. background, other, are handled differently by all four methods, so the same approach is used for each dataset in all four methods to ensure fair comparison. We modified the openly-available Tensorflow implementations of SEC\footnote{\url{https://github.com/xtudbxk/SEC-tensorflow}} and DSRG\footnote{\url{https://github.com/xtudbxk/DSRG-tensorflow}}, as well as the Keras implementation of HistoSegNet\footnote{\url{https://github.com/lyndonchan/hsn_v1}}. We have released the full evaluation code for this paper online\footnote{\url{https://github.com/lyndonchan/wsss-analysis}}.

For each dataset, the eight network-method configurations are quantitatively ranked against the ground-truth annotated evaluation sets using the mean Intersection-over-Union (mIoU) metric, which measures the percent overlap between predicted ($P$) and ground-truth segmentation masks ($T$), averaged across all $C$ classes (see Equation \ref{eqn_miou}). Qualitative evaluation is provided by visual inspection of the segmentation quality. Grad-CAM (using the most confident class at each pixel) is used as the baseline for both qualitative and quantitative evaluations (see Section 3 of the Supplementary Materials for details).

\begin{equation*}
	\mathrm{mIoU}=\frac{1}{C}\sum_{c=1}^{C}\frac{|P_c\cap T_c|}{|P_c\cup T_c|}\tag{1}
	\label{eqn_miou}
\end{equation*}


\subsection{Atlas of Digital Pathology (ADP)}

HistoSegNet was originally developed for ADP and hence needs no modifications, but SEC, DSRG, and IRNet were modified to generate \emph{background} and \emph{other} functional class activations by measuring the white level and the negative of the maximum of other functional classes. The foreground CAMs were thresholded at $90\%$ of the maximum value for SEC and DSRG, with the same overlap strategy used; thresholding at 0.9 was used for IRNet. SEC and DSRG were trained for 8 epochs, IRNet for 3 epochs. See Sections 4.3, 4.4 of the Supplementary Materials for the detailed settings and training progress of SEC and DSRG in ADP-morph and ADP-func respectively; see Section 5.3, and 5.4 for IRNet.

\textbf{Quantitative Performance}

\begin{figure}[h!]
	\centerline{
		\subfigure[Morphological types]{\includegraphics[width=\linewidth]{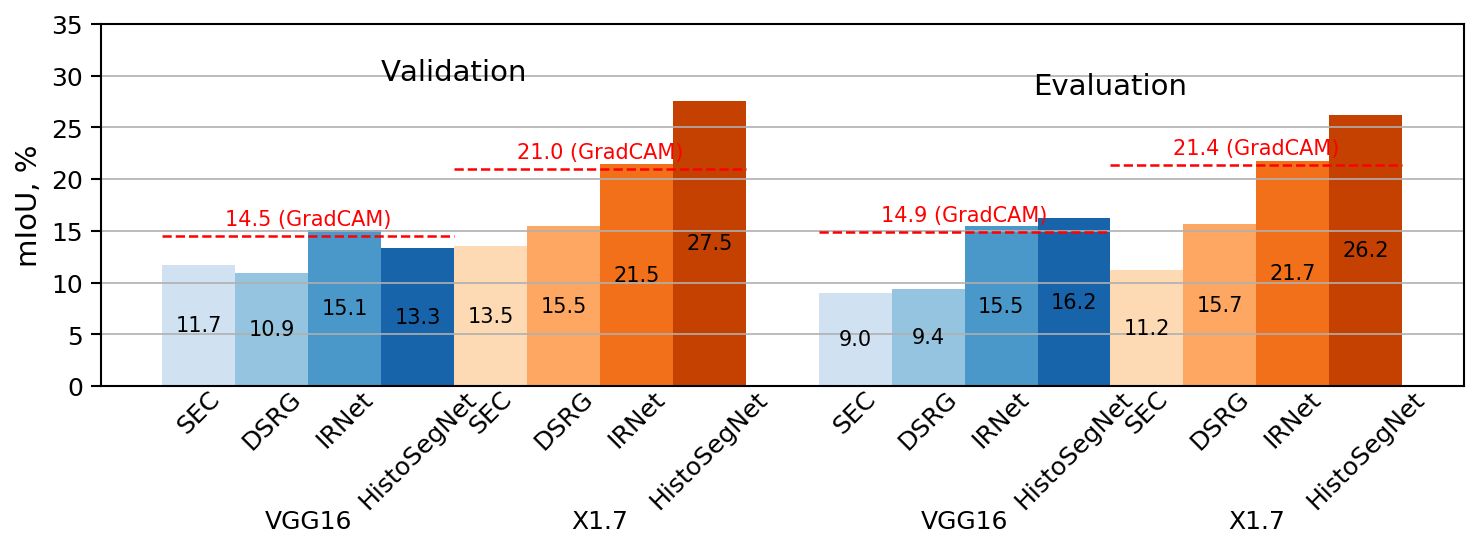}\label{fig_eval_ADP_morph}}
	}
	\centerline{
		\subfigure[Functional types]{\includegraphics[width=\linewidth]{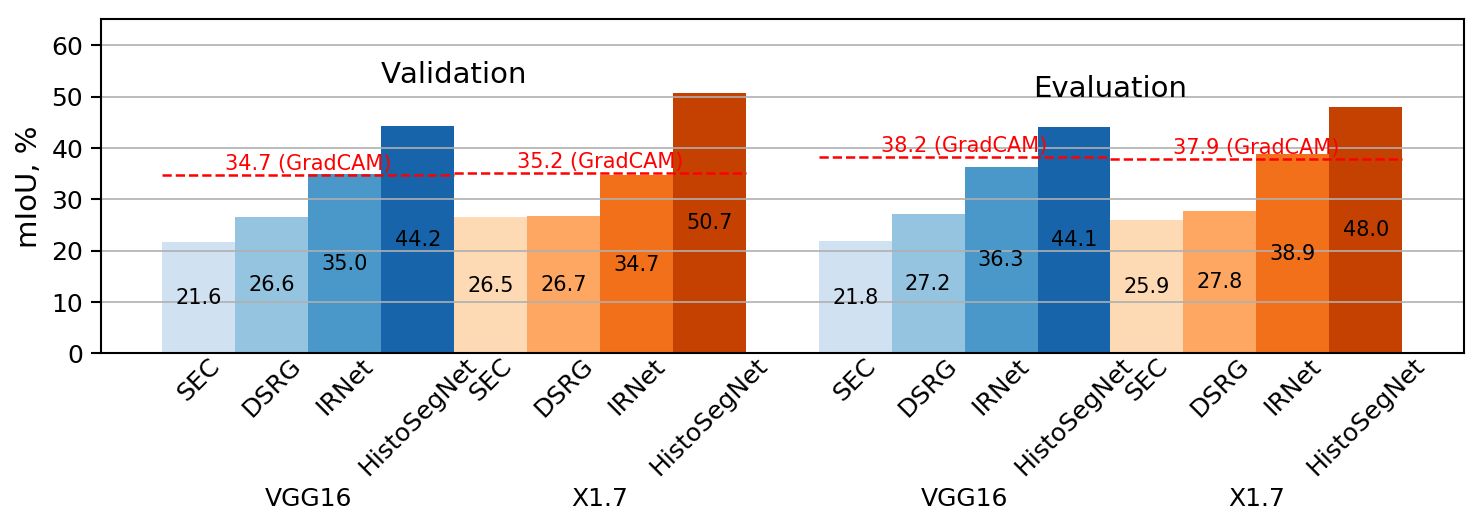}\label{fig_eval_ADP_func}}
	}
 \caption{ADP: quantitative performance of evaluated configurations (and baseline Grad-CAM) on the tuning (left) and evaluation set (right).} %
\label{fig_eval_ADP}
\end{figure}

When assessed against the ground-truth evaluation set for both morphological and functional types (see Figure \ref{fig_eval_ADP}), it may be seen that (1) HistoSegNet is the only method that consistently out-performs the baseline Grad-CAM and that (2) the X1.7 network (which was designed for ADP) is superior to VGG16. Among the self-supervised methods, SEC performs worst, followed by DSRG; IRNet performs as well as Grad-CAM. As HistoSegNet was tuned with the validation set, it performs somewhat worse on the evaluation set (which has fewer unique classes per image).

\textbf{Qualitative Performance}

\begin{figure}[h!]
	\centerline{
		\subfigure[Colour key]{\includegraphics[width=\linewidth]{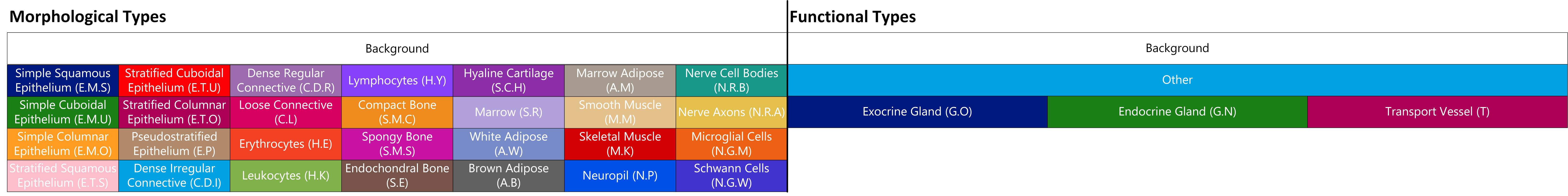}\label{fig_key_ADP}}
	}
	\centerline{
		\subfigure[Morphological types]{\includegraphics[width=\linewidth]{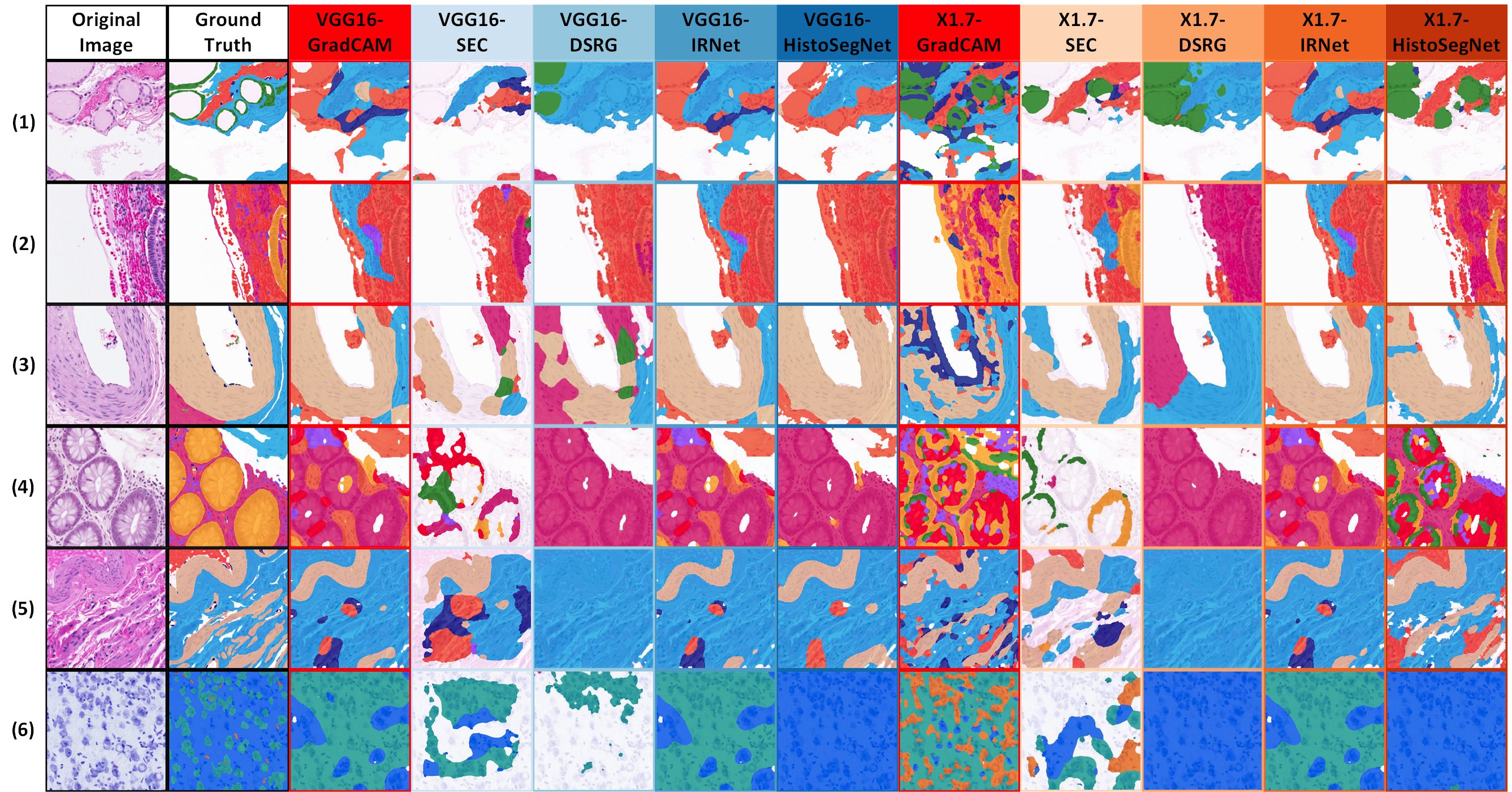}\label{fig_qual_ADP_morph}}
	}
	\centerline{
		\subfigure[Functional types]{\includegraphics[width=\linewidth]{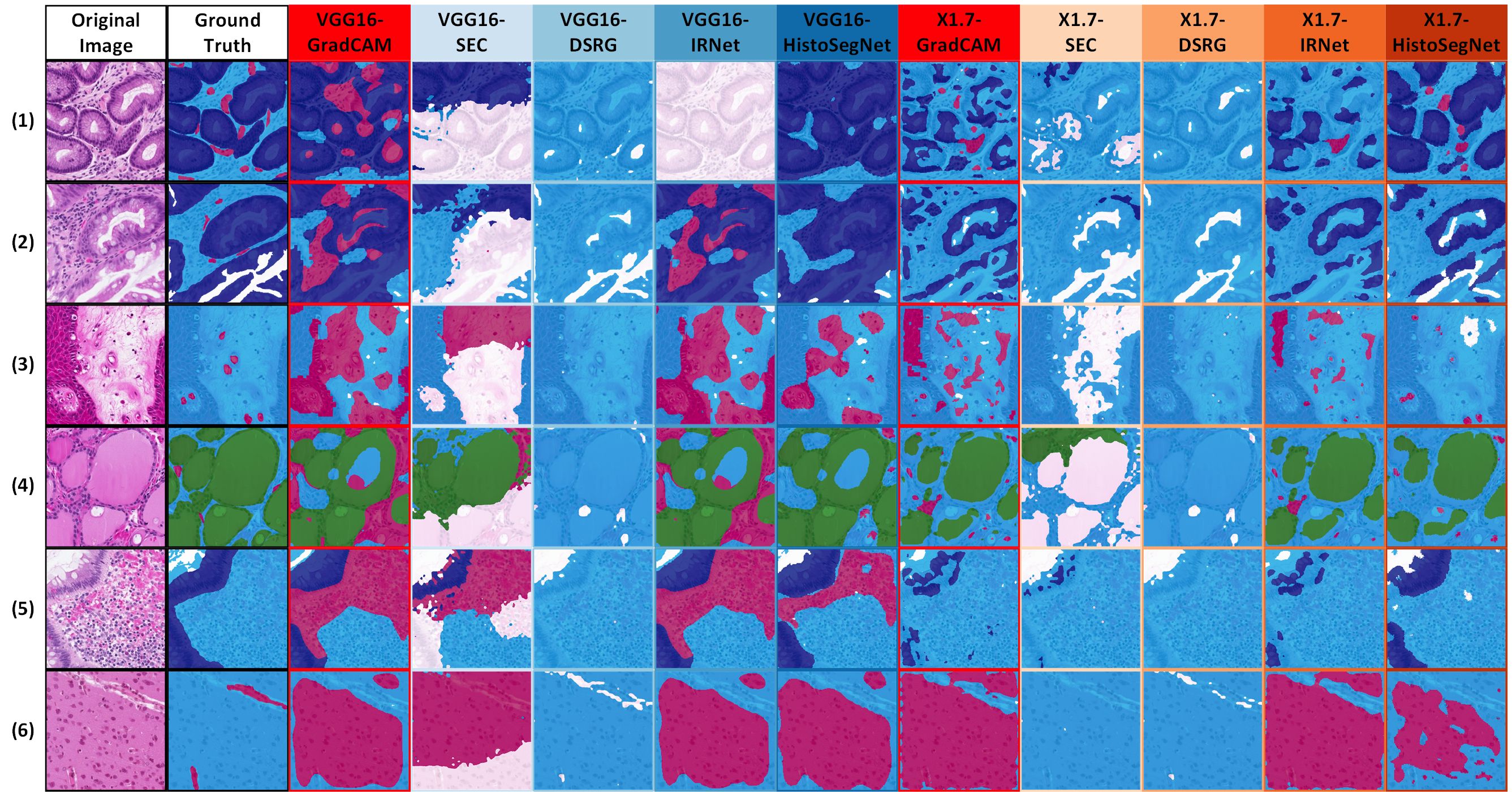}\label{fig_qual_ADP_func}}
	}
 \caption{ADP: qualitative performance of evaluated configurations (and baseline Grad-CAM), on select evaluation patch images.} %
\label{fig_qual_ADP}
\end{figure}

Figure \ref{fig_qual_ADP} visualizes the segmentation performances for select patches. For the morphological types (see Figure \ref{fig_qual_ADP_morph}), the X1.7 configurations are superior to the VGG16 configurations (since the X1.7 Grad-CAMs correspond better to the smaller segments). While SEC and DSRG correspond well with object contours, they tend to over-exaggerate object sizes whereas HistoSegNet does not. For example, in image (1) of \ref{fig_qual_ADP_morph}, only X1.7-HistoSegNet accurately segments the \emph{simple cuboidal epithelium} of the thyroid glands (in green), although it struggles to delineate the \emph{lymphocytes} (purple) in image (4) and \emph{neuropil} (blue) in image (6). Similar behaviour is observed for the functional types (see Figure \ref{fig_qual_ADP_func}): in images (1)-(4), only HistoSegNet detects small transport vessels (in fuchsia) although it produces false positives in images (5)-(6).

\subsection{PASCAL VOC2012}

SEC, DSRG, and IRNet were originally developed for PASCAL VOC2012 but new seeds were generated using our experimental framework and utilized for all methods. We used the \emph{background} activation from SEC (i.e. the negative class sum of CAMs from the ``background'' network) for all four methods (including HistoSegNet) since the DRFI \citep{jiang2013salient} from DSRG had no readily implementable code and HistoSegNet's white-illumination assumption is not applicable here. Both SEC and DSRG were trained for 16 epochs, IRNet for 3 epochs. See Sections 4.5 and 5.5 of the Supplementary Materials for the detailed settings and training progress of SEC and DSRG, and IRNet respectively.

\textbf{Quantitative Performance}

\begin{figure}[h!]
	\begin{center}
			 \includegraphics[width=.7\linewidth]{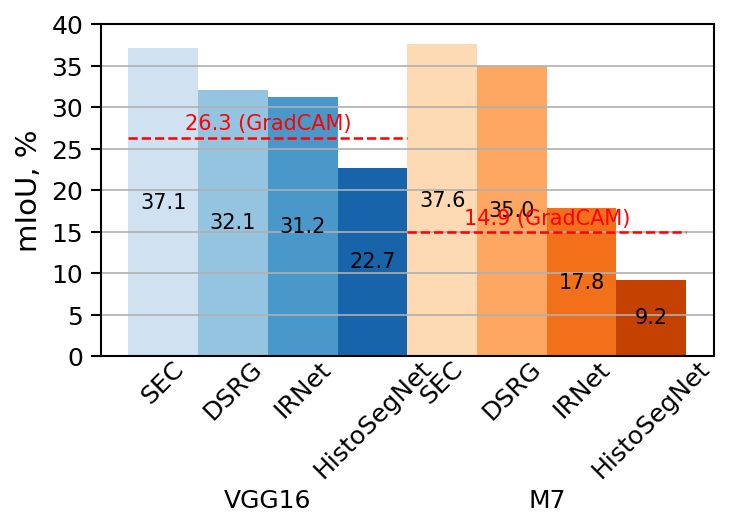}
	\end{center}
   \caption{PASCAL VOC2012: quantitative performance of evaluated network-method configurations (and baseline Grad-CAM) on the evaluation set.}
\label{fig_eval_VOC2012}
\end{figure}

When assessed against the ground-truth evaluation set (see Figure \ref{fig_eval_VOC2012}), it may be seen that (1) only SEC and DSRG consistently out-perform the baseline Grad-CAM, with SEC being clearly superior and that (2) the VGG16 network is overall superior to the M7 network. SEC using M7 cues performs the best overall (slightly better than SEC with VGG16 cues). Furthermore, we obtained results for SEC, DSRG, and IRNet somewhat inferior to those originally reported. We suspect that (1) neglecting to use DRFI for background cue generation in DSRG and (2) minor implementation differences between the Caffe and TensorFlow implementations are responsible for this in SEC and DSRG, since we observed discrepancies between our generated cues and those provided by the authors. For IRNet, using square image resizing and shallower networks than ResNet-50 may have caused decreased performance.

\textbf{Qualitative Performance}

\begin{figure}[h!]
	\centerline{
		\subfigure[Colour key]{\includegraphics[width=\linewidth]{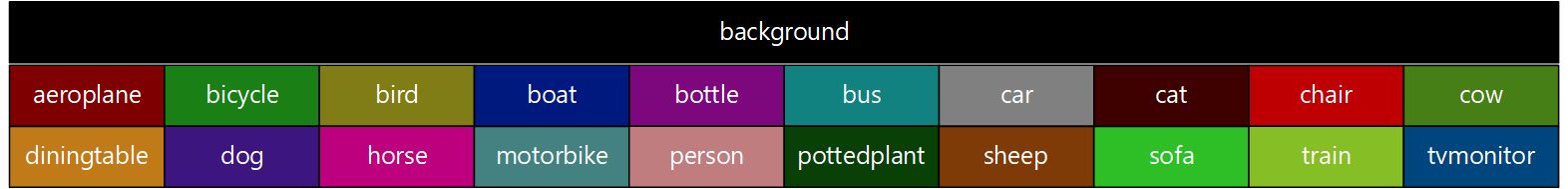}}
	}
	\centerline{
		\subfigure[Segmentation results]{\includegraphics[width=\linewidth]{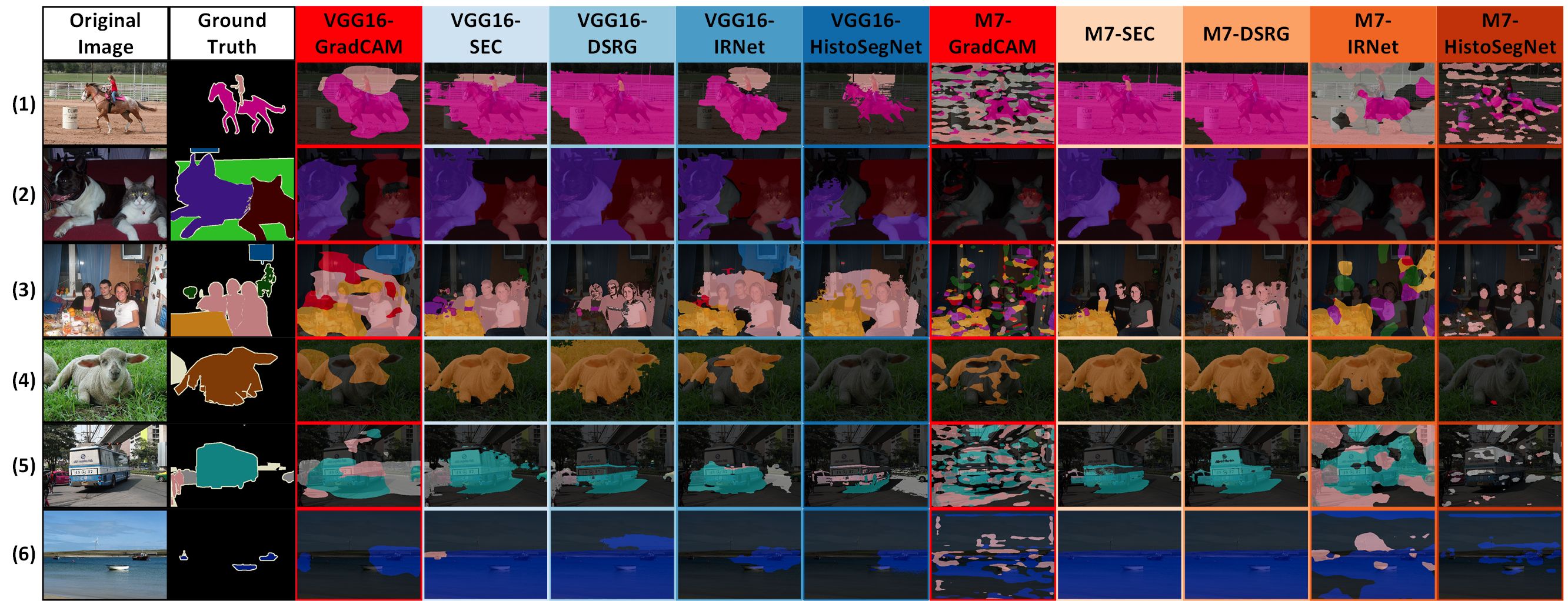}}
	}
	\caption{PASCAL VOC2012: qualitative performance of evaluated configurations (and baseline Grad-CAM), on select evaluation images}
\label{fig_qual_VOC2012}
\end{figure}

Figure \ref{fig_qual_VOC2012} visualizes each configuration's segmentation results for several representative images. It is evident that the VGG16 Grad-CAM captures entire objects while the M7 Grad-CAM only captures parts and this results in the VGG16 configurations performing better. Furthermore, SEC and DSRG are able to correct mistakes in the original Grad-CAM (possibly due to the seeding loss function being well-suited to this dataset) whereas HistoSegNet often connects Grad-CAM segments to the wrong objects. In image (3), VGG16-HistoSegNet confuses the \emph{diningtable} segment (yellow) with \emph{person} segments (peach) while M7-HistoSegNet only segments heads and arms as \emph{person}. All methods struggle most to differentiate objects that frequently occur together, such as \emph{boat} and water in image (6).

\subsection{DeepGlobe Land Cover Classification}

The DeepGlobe Land Cover Classification dataset was intended for fully-supervised semantic segmentation, so no published WSSS has ever been developed for it. We ignore the extremely uncommon \emph{unknown} class for non-land cover objects, so all four methods consider the six land cover classes to be foreground. SEC and DSRG were trained for 13 epochs, IRNet for 3 epochs. See Sections 4.6 and 5.6 of the Supplementary Materials for the detailed settings and training progress of SEC and DSRG, and IRNet respectively.

\textbf{Quantitative Performance}

\begin{figure}[h!]
	\begin{center}
			 \includegraphics[width=.7\linewidth]{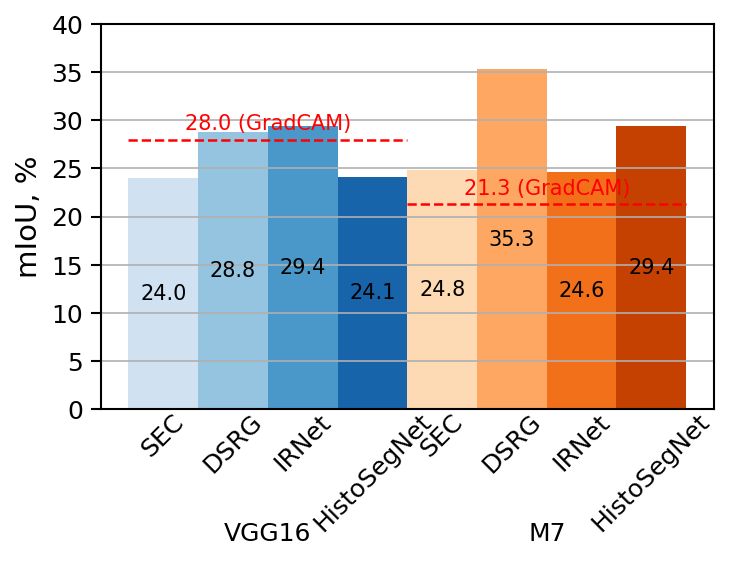}
	\end{center}
   \caption{DeepGlobe: quantitative performance of evaluated configurations (and baseline Grad-CAM) on the evaluation set.}
\label{fig_eval_DeepGlobe}
\end{figure}

When assessed against the ground-truth evaluation set (see Figure \ref{fig_eval_DeepGlobe}), (1) only DSRG and IRNet consistently out-perform the baseline Grad-CAM, although DSRG appears superior in general and (2) the M7 network is superior to VGG16, despite M7's Grad-CAM being inferior. DSRG using M7 cues performs the best overall. None of the methods were developed for DeepGlobe and the best fully-supervised method, DFCNet \citep{tian2019dense}, which attained a 52.24\% mIoU on the unreleased validation set, performs far better. Nonetheless, DSRG, IRNet, and HistoSegNet all out-perform the baseline while SEC does not.

\textbf{Qualitative Performance}

\begin{figure}[h!]
	\centerline{
		\subfigure[Colour key]{\includegraphics[width=\linewidth]{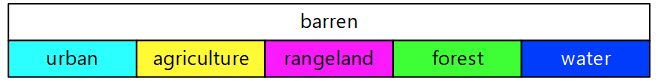}}
	}
	\centerline{
		\subfigure[Segmentation results]{\includegraphics[width=\linewidth]{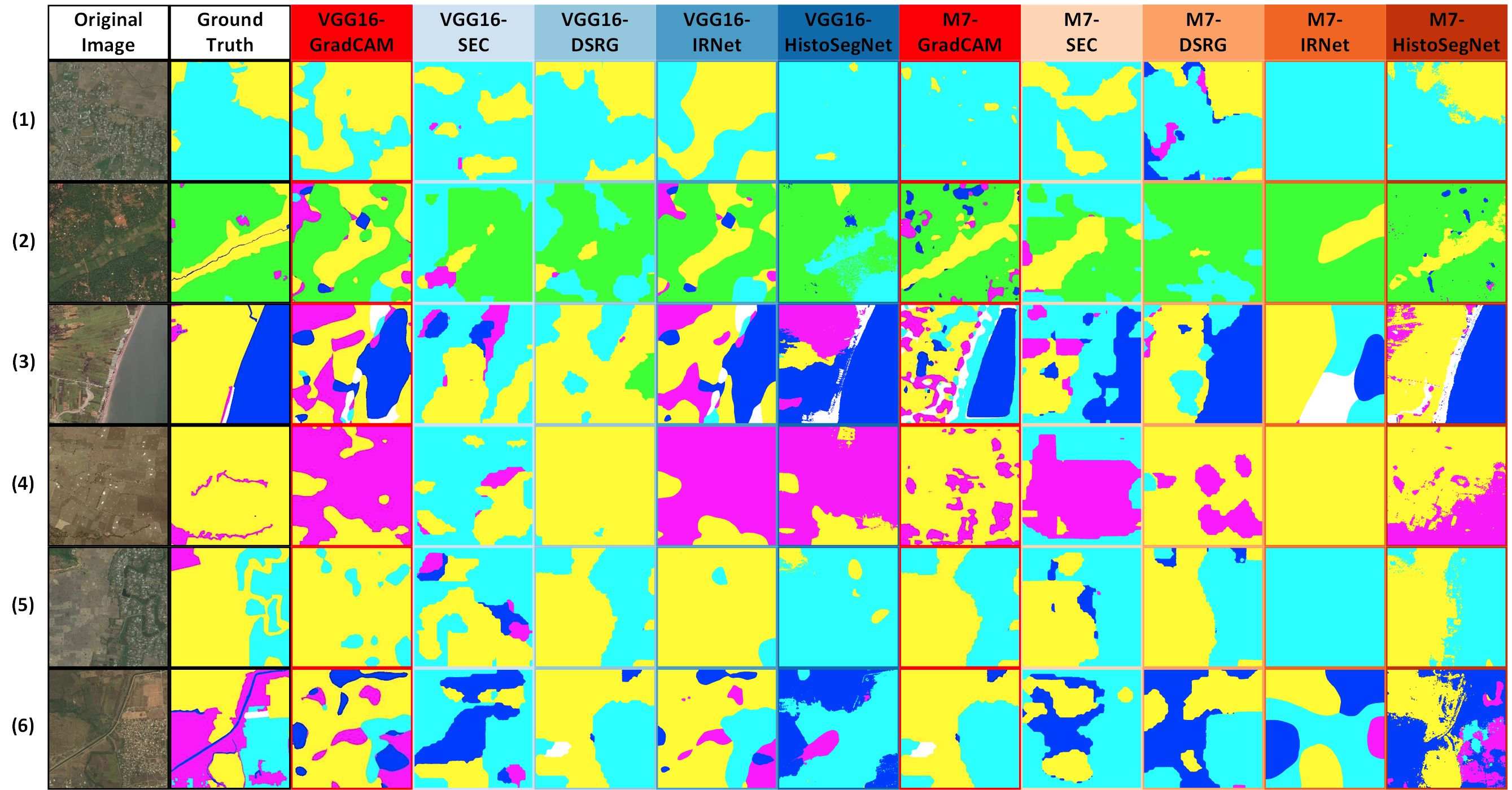}}
	}
	\caption{DeepGlobe: qualitative performance of evaluated configurations (and baseline Grad-CAM), on select evaluation images.}
\label{fig_qual_DeepGlobe}
\end{figure}

Figure \ref{fig_qual_DeepGlobe} displays the segmentation results for several images. Visually, all four methods predict rather similarly, although DSRG and HistoSegNet capture small details better, and VGG16 methods tend to produce coarser predictions than M7. Unlike in VOC2012, the Grad-CAMs already capture the rough locations of the segments accurately, and only minor modifications are needed. For example, the M7 Grad-CAMs successfully detect the \emph{agriculture} segment (yellow) in the middle of image (2) and the \emph{rangeland} (magenta) in the bottom of image (4) but only HistoSegNet retains these preliminary segments. All methods struggle with segmenting \emph{water} (blue), however, as shown in image (6).

\section{Analysis}\label{sec_analysis}

Since the same \textcolor{mygreen}{four} WSSS methods were compared with identical classification networks (or the closest equivalents) with the same evaluation setup in three datasets, it is possible to compare their comparative suitability for each dataset and observe some common themes. This is crucial, since WSSS in other image domains than natural scene images and histopathology images has been largely unexplored and applying them to these image domains requires an understanding of which approaches are best suited to the dataset at hand even before training. In this section, we analyze (1) the effect of the sparseness of classification network cues, (2) whether self-supervised learning is beneficial, and (3) how to address high class co-occurrence in the training set.

\subsection{Effect of Classification Net Cue Sparseness}\label{sec_a1}

\begin{figure}[h!]
	\begin{center}
			 \includegraphics[width=.6\linewidth]{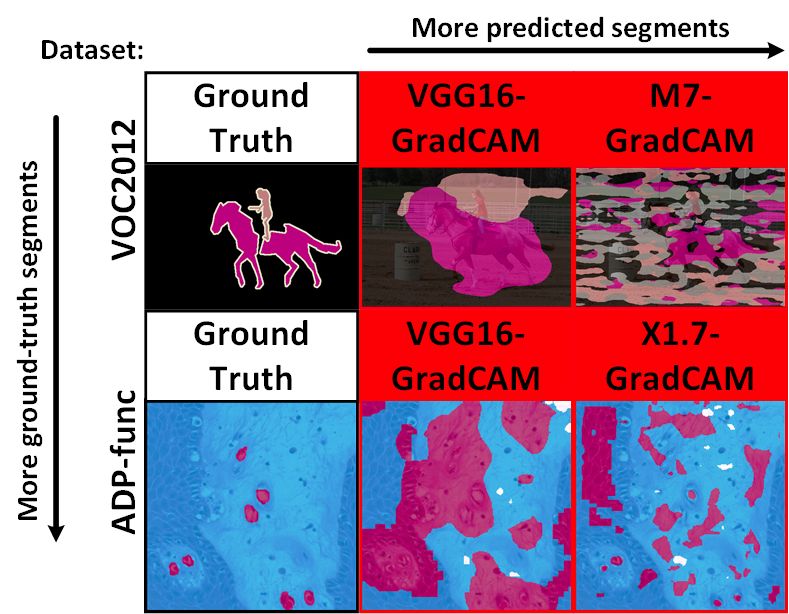}
	\end{center}\vspace{-.1in}
   \caption{Networks with more predicted segments (i.e. X1.7/M7) perform better than those with fewer (VGG16) on datasets with more segments (ADP-func) than fewer (VGG16).}
	\label{fig_a1_qual}
\end{figure}

In most WSSS methods, not much attention is paid to the design of the classification network used. SEC and DSRG use VGG16 and HistoSegNet uses X1.7 (or M7). However, our experimental results showed that the choice of classification network has a significant effect on subsequent WSSS performance. Heuristically, we observed that networks generating sparser Grad-CAM segments would also perform better on datasets with more ground-truth segments. This was true for both the baseline Grad-CAMs and also subsequent WSSS performance. In Figure \ref{fig_a1_qual}, this is demonstrated using a sample image from VOC2012 and ADP-func: the selected VOC2012 image has three ground-truth segments, while the ADP-func image has eight. VGG16's Grad-CAM predicts fewer segments because its final feature map is sized $41\times 41$ (with input size of $321\times 321$), but predicts sparser cues with M7 (and X1.7) because its final feature map is $56\times 56$ (with input size of $224\times 224$) and hence requires less upsampling. While VGG16 captures the spatial extent of the \emph{person} and \emph{horse} better than M7 in VOC2012, it is too coarse for ADP-func and X1.7 performs better.

\begin{figure}[h!]
	\begin{center}
			 \includegraphics[width=\linewidth]{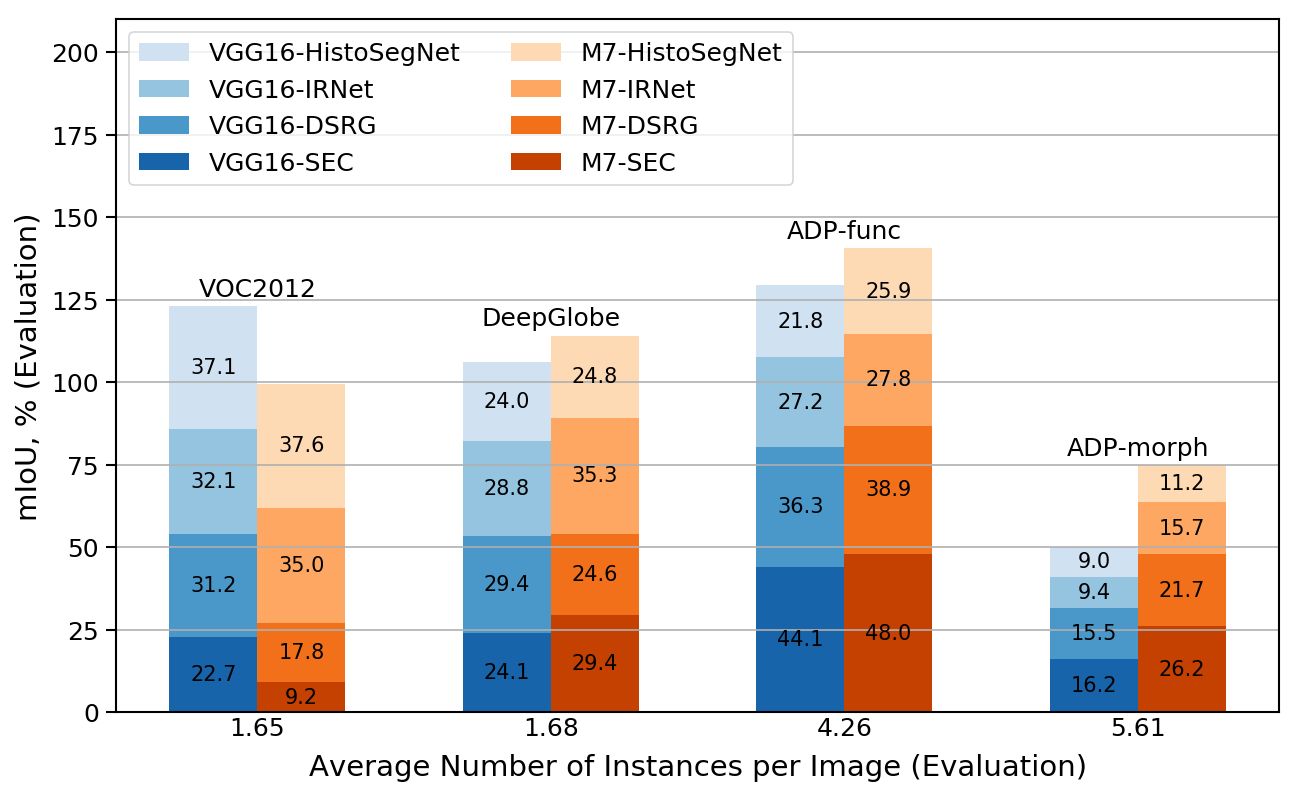}
	\end{center}\vspace{-.1in}
   \caption{Datasets with more ground-truth instances tend to be segmented better by X1.7/M7 (which has a larger feature map) than VGG16 (which has a smaller feature map).}
	\label{fig_a1}
\end{figure}

This heuristic observation is also confirmed by quantitative analysis of the relation between the number of ground-truth instances and segmentation performance in the evaluation set. In Figure \ref{fig_a1}, the evaluation set mIoU of each configuration is shown for the three datasets after ordering by increasing number of ground-truth instances. VGG16 configurations (in shades of blue) perform best in datasets with fewer ground-truth instances ($\leq$1.65), while M7 configurations (in shades of orange) perform best in datasets with more ground-truth instances ($\geq$1.68). The effect is not insignificant, causing a mean difference of 5.22\% in mIoU, and is especially pronounced for HistoSegNet. These results suggest that it is worthwhile to select a classification network with appropriately sparse cues for each new dataset based on the number of ground-truth instances.

\subsection{Is Self-Supervised Learning Beneficial?}\label{sec_a2}

The prevailing approach to WSSS is currently to generate weak cues using CAM or Grad-CAM for self-supervised learning of an FCN (as used by SEC and DSRG). While this approach works well for natural scene images, our experimental results showed that it is clearly inferior for histopathology and of dubious value for satellite images. Why does self-supervised learning work well for some images and not others? Is it possible to determine ahead of time which approach is suitable for a given dataset before training? Heuristically, it was observed that self-supervised learning performance was heavily dependent on the degree to which ground-truth segments were already covered by the thresholded Grad-CAM seeds (adjusted to cover just under 50\% of the image). In Figure \ref{fig_qual_VOC2012}, a sample image and the associated M7/X1.7 Grad-CAM segmentation is shown from VOC2012 and ADP-func respectively. The M7/X1.7 cue covers very little of the ground-truth \emph{sheep} (tan-coloured) in the VOC2012 image but covers almost the entire ground-truth \emph{other} in the ADP-func image; the self-supervised methods (SEC, DSRG, and IRNet) subsequently segment the VOC2012 image better while HistoSegNet (which is not self-supervised) segments the ADP-func image better.

\begin{figure}[h!]
	\begin{center}
			 \includegraphics[width=\linewidth]{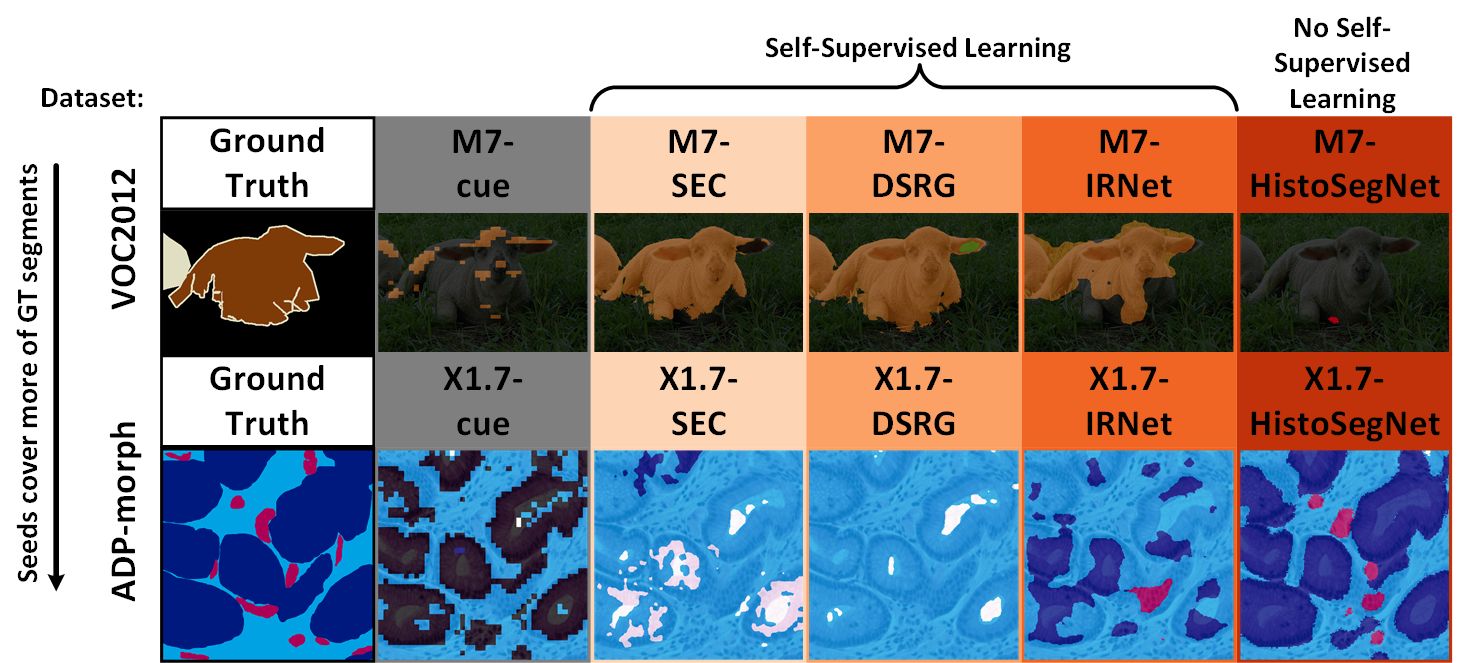}
	\end{center}\vspace{-.1in}
   \caption{Methods without self-supervised learning (i.e. HistoSegNet) perform better on datasets where seeds already cover much of the ground-truth segments (i.e. ADP-func), known as mean recall. Self-supervised learning seems to be a poor choice in these cases.}
\label{fig_a2_qual}
\end{figure}

Quantitatively, the degree of overlap between predicted seeds and ground-truth segments in the evaluation set can be measured by mean recall, which is the percent of ground-truth pixels $T_c$ that are predicted $P_c$ correctly for each class $c$, then averaged across all $C$ classes (see Equation \ref{eqn_re}):

\begin{equation*}
	\mathrm{Recall}=\frac{1}{C}\sum_{c=1}^{C}\frac{|P_c\cap T_c|}{|T_c|}\tag{2}
	\label{eqn_re}
\end{equation*}

\begin{figure}[h!]
	\begin{center}
			 \includegraphics[width=\linewidth]{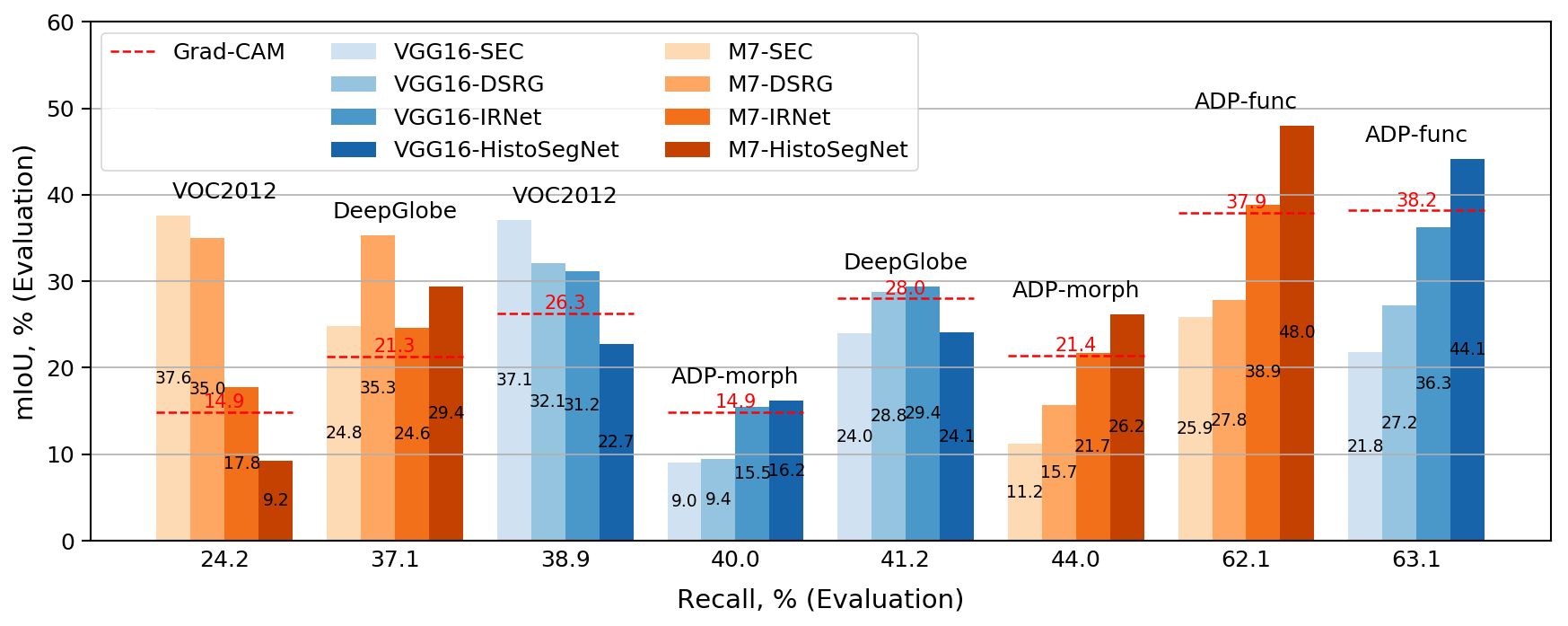}
	\end{center}\vspace{-.1in}
   \caption{Datasets with higher mean seed recall (after thresholding below 50\% seed coverage) are overwhelmingly better segmented by non-self supervised learning methods (i.e. HistoSegNet). Where recall is low, self-supervised methods (i.e. SEC, DSRG, IRNet) are superior.}
	\label{fig_a2}
\end{figure}

In Figure \ref{fig_a2}, the evaluation set mIoU of each configuration is shown for each dataset's cues after ordering by increasing mean recall. Self-supervised methods (SEC, DSRG, and IRNet) tend to perform better than the cues for datasets with low seed coverage (such as VOC2012 and DeepGlobe) but HistoSegNet performs better for datasets with high seed coverage (ADP-func and ADP-morph). This suggests that, when applying WSSS to a new dataset, one should choose a self-supervised method (e.g. SEC and DSRG) if seed recall is low ($<40\%$) and a method without self-supervised learning (e.g. HistoSegNet) if seed recall is high ($\geq40\%$). This makes much intuitive sense, since CAM/Grad-CAM is notorious for only segmenting parts of ground-truth objects in VOC2012. Hence, self-supervised methods were developed with loss functions to encourage predicting liberally from minimal seeds by rewarding true positives and not penalizing false positives. While this strategy works well when seeds cover little of the ground-truth, it is clearly detrimental when the seeds cover much of the ground-truth.

\begin{figure}[h!]
	\centerline{
		\subfigure[SEC/DSRG mIoU.]{\includegraphics[width=0.5\linewidth]{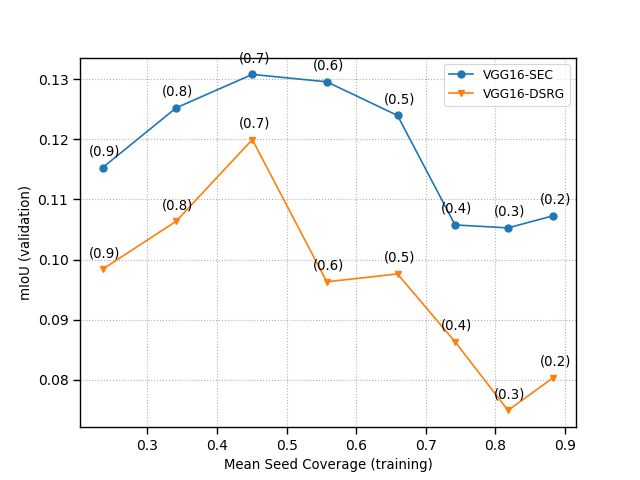}\label{fig_a2_thresh_miou}}
		\subfigure[VGG16 seed precision/recall]{\includegraphics[width=0.5\linewidth]{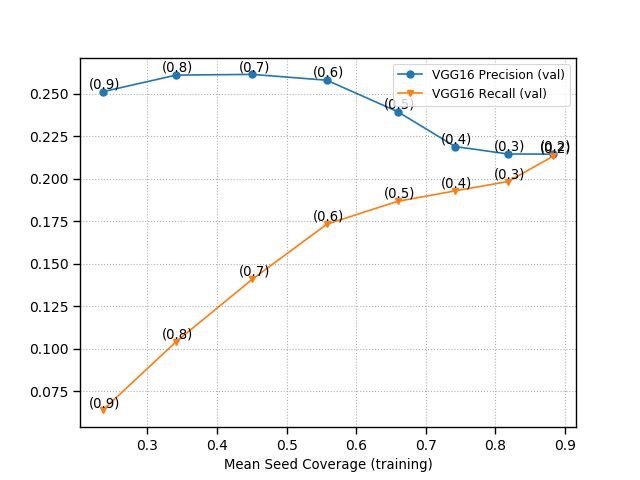}\label{fig_a2_thresh_pr_re}}
	}
   \caption{Mean seed coverage in the training set (threshold levels in brackets), plotted versus mIoU of VGG16-SEC and VGG16-DSRG for ADP-morph validation set (left), versus seed precision/recall (right). Although mean recall is maximized by increasing seed coverage, SEC and DSRG perform best when seed coverage is fixed at just below 50\%.}
	\label{fig_a2_thresh}
\end{figure}

\newpage
Although the previous analysis determined that high mean recall in the week seeds is beneficial for SEC and DSRG, lowering the seed threshold to ensure more than 50\% of the image area is seeded (mean seed coverage) and thus increase the mean recall is not a feasible strategy either. In Figure \ref{fig_a2_thresh_miou}, a simple ablative study is shown of the relationship between mIoU in ADP-morph for VGG16-SEC and VGG16-DSRG with different seed threshold levels (i.e. 20\% to 90\%). Figure \ref{fig_a2_thresh_pr_re} shows the effect of these same threshold levels on seed precision and recall. Although decreasing the seed threshold level to 20\% (and increasing seed coverage to 88.3\%) increases mean recall to 21.3\%, the optimal seed threshold level for SEC and DSRG mIoU is actually when seed coverage is just below 50\%, resulting in a lower mean recall of 19.3\%. This simple analysis indicates that self-supervised methods such as SEC and DSRG are inherently ill-suited for datasets with low seed recall; the seed threshold level should be fixed so seed coverage is just below 50\% and decreasing the threshold to increase recall cannot improve performance. Full details are available in Section 9 of the Supplementary Materials.


\subsection{Addressing High Class Co-Occurrence}\label{sec_a3}

\begin{figure}[h!]
	\centerline{
		\subfigure[Training set, without balancing (75\% train)]{\includegraphics[width=0.5\linewidth]{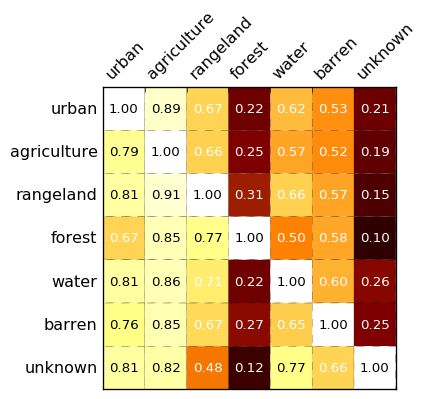}\label{coocc_DeepGlobe_nobalance}}
		\subfigure[Training set, with balancing (37.5\% train)]{\includegraphics[width=0.5\linewidth]{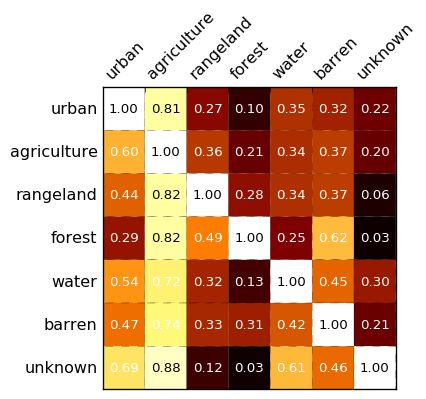}\label{coocc_DeepGlobe_balanced}}
	}
	\caption{Normalized class co-occurrences in the ground-truth image annotations of different DeepGlobe train-test splits. By removing the training images with the most annotated classes (i.e. balancing), training set class co-occurrence is significantly reduced in the \emph{rangeland}, \emph{forest}, and \emph{water} classes.}
\label{fig_coocc_DeepGlobe}
\end{figure}

Learning semantic segmentation from image labels is the weakest form of supervision possible because it provides no location information of the objects. This information must be inferred by their presence or absence in the annotated images. Logically, it would make sense that image label supervision would be least informative in datasets where the classes frequently occur together. In the extreme case that two labels always occur together, it would be impossible to learn to spatially separate them. The DeepGlobe dataset, for example, has very high levels of class co-occurrence (see Figure \ref{fig_coocc_DeepGlobe}) - the classes in the original training set (see Figure \ref{coocc_DeepGlobe_nobalance}) regularly co-occur in more than 50\% of images (except for \emph{forest} and \emph{unknown}). To assess whether simply reducing class co-occurrence would improve WSSS performance, we removed half of these original training images with the most class labels (defined as the sum of overall class counts for each image) and then retrained - we call this process ``balancing'' the class distribution. As a result, the class co-occurrence is significantly reduced (see Figure \ref{coocc_DeepGlobe_balanced}) in all classes except \emph{urban} and \emph{agriculture}.

\begin{figure}[h!]
	\begin{center}
			 \includegraphics[width=\linewidth]{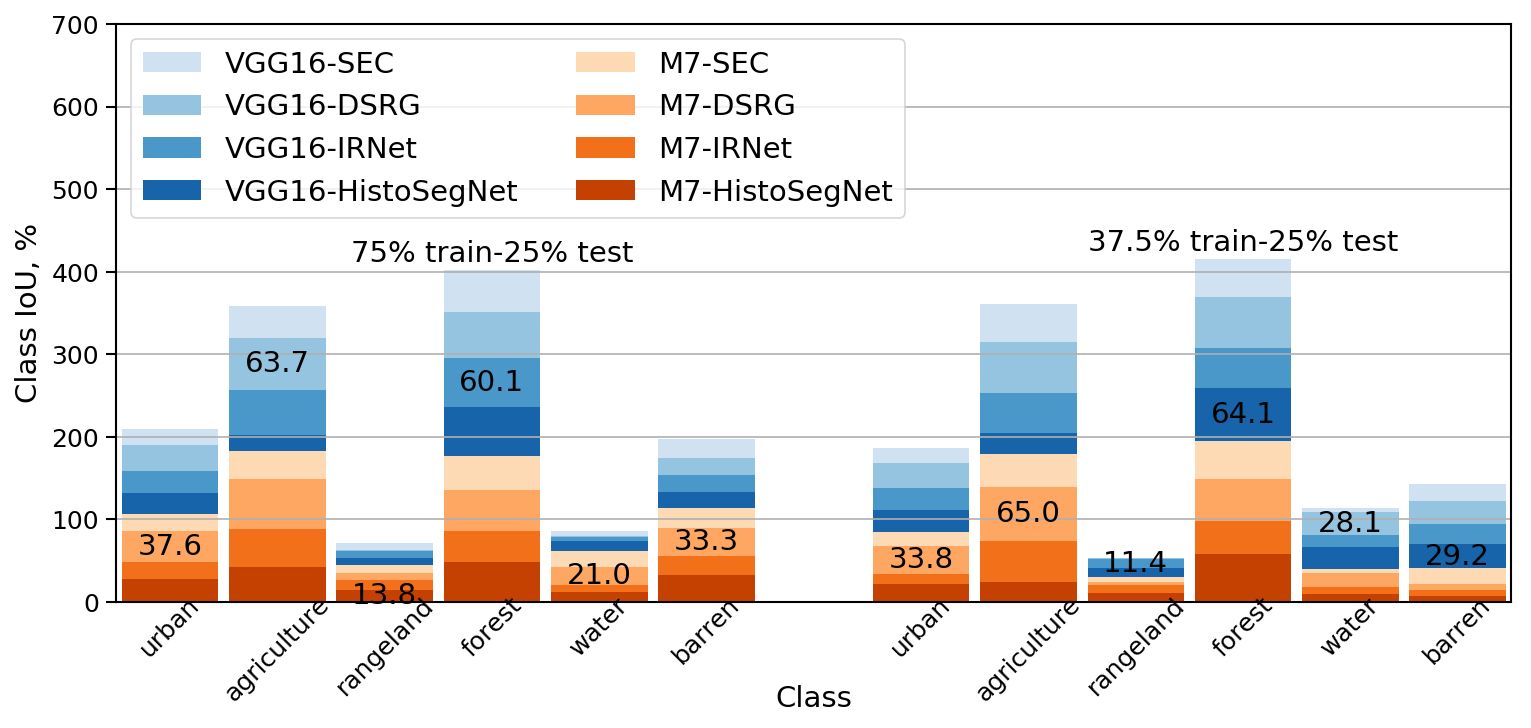}
	\end{center}\vspace{-.1in}
   \caption{Class IoU of the best-performing configuration increases for the \emph{agriculture}, \emph{forest}, and \emph{water} after balancing (right), compared to before balancing (left). Some classes experience a performance decrease, suggesting that the balancing method has room for improvement.}
\label{fig_a3}
\end{figure}

When we use these two different train-test splits and evaluate on the same test set, we obtain the quantitative mIoU performances shown in Figure \ref{fig_a3} - results before balancing are shown on the left and after balancing on the right. Although performance in certain methods deteriorate significantly after balancing, the best performance improves in the classes that experienced the greatest changes from the balancing, such as \emph{agriculture}, \emph{forest}, and \emph{water}. See Sections 4.7 and 5.7 of the Supplementary Materials for the detailed training progress of SEC and DSRG, and IRNet respectively.

\begin{figure}[h!]
	\begin{center}
			 \includegraphics[width=.8\linewidth]{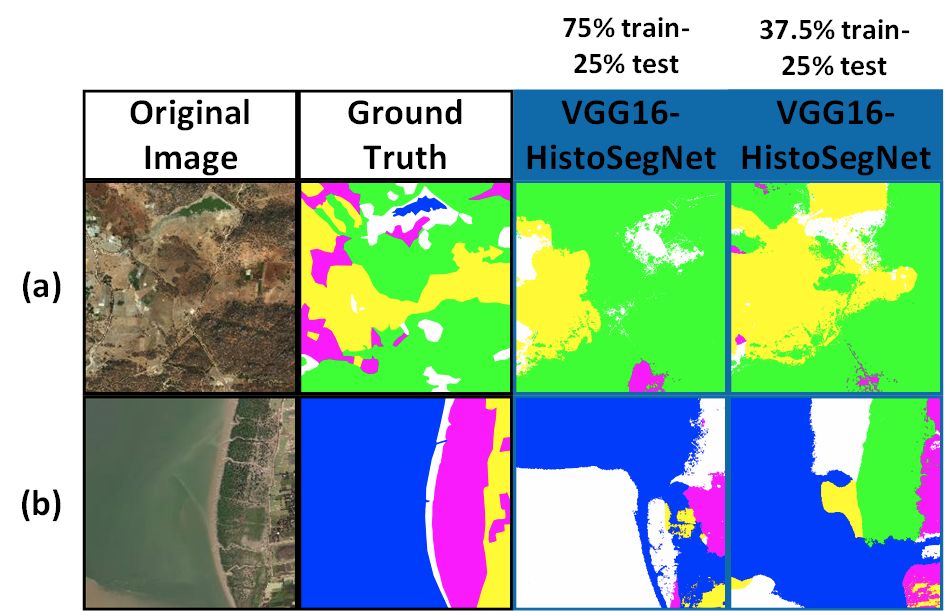}
	\end{center}\vspace{-.1in}
   \caption{Segmentation by VGG16-HistoSegNet for DeepGlobe, trained without (second from right) and with class balancing (right). Performance improves most in classes experiencing the greatest changes, such as \emph{forest} and \emph{water}.}
	\label{fig_a3_qual}
\end{figure}

\newpage
This is confirmed upon inspecting some segmentation results for images containing \emph{forest} and \emph{water}, as shown in Figure \ref{fig_a3_qual} for VGG16-HistoSegNet. Since balancing drastically reduces class co-occurrence in these two classes, VGG16-HistoSegNet learns to delineate \emph{forest} from \emph{agriculture} better in image (a) and associate \emph{water} with the river on the left of image (b). This shows that class co-occurrence is a significant challenge for WSSS from image labels but a simple technique to reduce it can help performance in the affected classes. Overall, the mean mIoU decreases from 27.5\% to 26.5\% after balancing, so we hypothesize that more effective methods of reducing class co-occurrence in the training set can more robustly improve WSSS performance.

\section{Conclusion}\label{sec_conclusion}

Weak supervision with image labels is a promising approach to semantic segmentation (WSSS) because image annotations require significantly less expense and time than the pixel annotations needed for full supervision. To date, state-of-the-art WSSS methods have built their methodologies exclusively around natural scene images. However, the lack of methods built for alternative image domains indicates there is an implicit assumption that these methodologies are generalizable with minor modifications. However, while the major remaining challenges for natural scene images concern separating background from foreground and segmenting entire objects instead of parts, alternative image domains such as histopathology and satellite images present different challenges, such as ambiguous boundaries and class co-occurrence. This paper is the first to analyze whether state-of-the-art methods developed for natural scene images still perform acceptably on histopathology and satellite images and compares their performances against a method developed for histopathology images.

Our experiments indicated that state-of-the-art methods developed for natural scene (i.e. SEC and DSRG) and histopathology images (i.e. HistoSegNet) indeed performed best in their intended domains. Furthermore, we showed that most methods perform moderately well for satellite images. Many methods performed poorly on datasets they were not designed to solve (such as HistoSegNet in VOC2012, SEC and DSRG in ADP), although IRNet seemed to be most robust overal to dataset choice. We found that the sparseness of a classification network's baseline Grad-CAM had a significant effect on subsequent segmentation performance if the ground-truth segments were also sparse. We also observed that the self-supervised learning approach to WSSS was only beneficial if the optimally-seeded cues covered little of the ground-truth segments (low mean recall), and that methods forgoing the self-supervised learning approach performed better otherwise. Finally, we demonstrated the negative effect of class co-occurrence on segmentation performance and showed that even a simple method of reducing class co-occurrence can alleviate this problem.

The findings of our paper clearly indicate that mainstream methodologies are poorly suited for these other image domains. We believe that more work is needed to develop alternative methodologies for WSSS which are either specialized for these image domains or are at least more generalizable to them. Instead of the current focus on improving the recall of activation map seeds, perhaps devising new loss functions which refine ambiguous activation map boundaries or address high class co-occurrence would be better directions to explore in the future for histopathology and satellite images. Given the ease of collecting weak annotations and the inability of state-of-the-art algorithms to properly segment images from alternative image domains, the authors believe that it is imperative that more work be done to develop new methods capable of generalizing to different image domains.


%
%

\bibliographystyle{spbasic}      
\bibliography{Bibliography}   

\end{document}